\documentclass[lettersize,journal]{IEEEtran}
\usepackage[url = false, giveninits=true, backref=true, backrefstyle=two, block=space, natbib=true, backend=bibtex,  defernumbers=true, bibstyle=ieee, backref=true, citestyle=numeric-comp]{biblatex}
\usepackage{amsmath}
\usepackage{amsfonts}
\usepackage{amssymb}
\usepackage{algorithmic}
\usepackage{algorithm}
\usepackage{array}
\usepackage{textcomp}
\usepackage{stfloats}
\usepackage{url}
\usepackage{verbatim}
\usepackage{graphicx}
\usepackage{xcolor}
\usepackage[justification=centering,font={small,it},labelfont={small,bf},tableposition=top]{caption}
\usepackage{subcaption}
\usepackage{float}
\usepackage{multirow}
\usepackage{array}
\usepackage{tabularx}
\usepackage{colortbl}
\usepackage{xcolor}
\usepackage{booktabs}
\usepackage{multirow}
\usepackage{cuted}

\DeclareMathOperator*{\argmin}{argmin}

\begin{document}

\title{FH-DRL: Exponential–Hyperbolic Frontier Heuristics with DRL for Accelerated Exploration in Unknown Environments}

\providecommand{\AckUAM}{This research was partially supported by Basic Science Research Program through the National Research Foundation of Korea (NRF) funded by the Ministry of Education(No. 2020R1A6A1A03046811).}
\providecommand{\AckITRC}{This research was partially supported by the MSIT(Ministry of Science, ICT), Korea, under the ITRC(Information Technology Research Centre) support program(IITP-2020-2016-0-00465) supervised by the IITP(Institute for Information \& communications Technology Planning \& Evaluation).}
\providecommand{\AckMiddleSizedResearcher}{This research was supported by Basic Science Research Program through the National Research Foundation of Korea(NRF) funded by the Ministry of Education(2021R1A2C2094943)}
\providecommand{\AckMiddleSizedResearcherUAM}{This research was supported by Basic Science Research Program through the National Research Foundation of Korea(NRF) funded by the Ministry of Education (No. 2020R1A6A1A03046811) and (2021R1A2C2094943).}
\providecommand{\AckKIAT}{This paper was supported by Korea Institute for Advancement of Technology (KIAT) grant funded by the Korea Government(MOTIE) (P0020536, HRD Program for Industrial Innovation).}

\author{Seunghyeop Nam, Tuan Anh Nguyen ~\IEEEmembership{IEEE Member}, Eunmi Choi, Dugki Min
	\thanks{Seunghyeop Nam is with Department of Computer Science and Engieering, Konkuk University, Seoul 05029, South Korea, tomska@konkuk.ac.kr}%
	\thanks{Tuan Anh Nguyen and Dugki Min are with the Department of Artificial Intelligence, Graduate School, Konkuk University, Seoul 05029, South Korea, {anhnt2407, dkmin}@konkuk.ac.kr}%
	\thanks{Eunmi Choi is the School of Software, College of Computer Science, Kookmin University, Seoul 02707, South Korea, {emchoi@kookmin.ac.kr}}
}
	\markboth{~Vol.~X, No.~X, X~X}%
{Nam \MakeLowercase{\textit{et al.}}}%



\maketitle

\begin{abstract}
	Autonomous robot exploration in large-scale or cluttered environments remains a central challenge in intelligent vehicle applications, where partial or absent prior maps constrain reliable navigation. This paper introduces \textit{FH-DRL}, a novel framework that integrates a customizable heuristic function for frontier detection with a Twin Delayed DDPG (TD3) agent for continuous, high-speed local navigation. The proposed heuristic relies on an exponential–hyperbolic distance score, which balances immediate proximity against long-range exploration gains, and an occupancy-based stochastic measure, accounting for environmental openness and obstacle densities in real time. By ranking frontiers using these adaptive metrics, \textit{FH-DRL} targets highly informative yet tractable waypoints, thereby minimizing redundant paths and total exploration time. We thoroughly evaluate \textit{FH-DRL} across multiple simulated and real-world scenarios, demonstrating clear improvements in travel distance and completion time over frontier-only or purely DRL-based exploration. In structured corridor layouts and maze-like topologies, our architecture consistently outperforms standard methods such as Nearest Frontier, Cognet Frontier Exploration, and Goal Driven Autonomous Exploration. Real-world tests with a Turtlebot3 platform further confirm robust adaptation to previously unseen or cluttered indoor spaces. The results highlight \textit{FH-DRL} as an efficient and generalizable approach for frontier-based exploration in large or partially known environments, offering a promising direction for various autonomous driving, industrial, and service robotics tasks.
\end{abstract}

\begin{IEEEkeywords}
	Frontier-base Exploration, Deep Reinforcement Learning, Heuristic Optimisation, Self Navigation, SLAM
\end{IEEEkeywords}

\section{Introduction}
Autonomous exploration in complex, uncharted environments remains a core challenge for intelligent vehicles and mobile robotics \cite{Sun2024, Niroui2019, Fan2023}. Despite recent advances in perception and navigation, many robotic systems still rely on partial or pre-existing maps, limiting their performance when operating in large-scale or dynamic settings. A fundamental concept to tackle this gap is \emph{frontier-based exploration}, wherein the robot incrementally discovers environment boundaries, or ``frontiers,'' between known and unknown regions \cite{Yamauchi1997}. By continually moving toward these boundaries, frontier-based methods can, in principle, guarantee full coverage of the space. However, purely heuristic frontier selection often leads to suboptimal or redundant paths, especially in cluttered, multi-frontier situations \cite{Lubanco2020, Cimurs2022}. Consequently, improvements are needed to address such inefficiencies, especially for highly dynamic or large-scale environments.

Recent developments in \emph{deep reinforcement learning} (DRL) suggest a potential remedy, enabling robots to autonomously learn navigation policies through interactions with the environment \cite{Peake2021, Cao2024}. In principle, DRL-based exploration agents can surpass human-engineered heuristics by dynamically adapting to varied and unpredictable conditions. Nevertheless, purely end-to-end DRL frameworks—where the robot learns everything from raw sensor streams to action outputs—are prone to several drawbacks: (1) the search space in large unknown environments is excessively high-dimensional; (2) the training process can be extremely data-intensive and unstable; and (3) guaranteeing full coverage with end-to-end policies is challenging unless carefully designed reward mechanisms and exploration strategies are in place \cite{Niroui2019, Cimurs2022}. Furthermore, standard robotic navigation stacks in ROS, such as \texttt{nav2}, generally depend on costmaps generated from partial prior knowledge and may prove inadequate in fully unknown areas.

\subsection{Motivation and Research Gap}
Bridging the strengths of frontier-based exploration with the adaptability of DRL presents an appealing hybrid direction. Traditional frontier algorithms robustly detect unknown regions but may not effectively optimize travel distance or completion time; DRL, conversely, learns to refine movement decisions but faces difficulties when searching the entire action space in large environments. Several works have investigated this synergy, typically by letting a learning agent select frontiers or vantage points, while classical motion-planning algorithms handle low-level control \cite{Xu2022, Lubanco2020}. Despite progress, key research gaps persist:
\begin{itemize}
	\item \textbf{Suboptimal frontier scoring}: The conventional reliance on static, distance-oriented heuristics can yield frequent revisits and overly conservative paths \cite{Yamauchi1997}. Balancing short-range distance efficiency with a frontier's long-term information gain remains a challenge.
	\item \textbf{Lack of an adaptable, integrated framework}: Many existing solutions fix certain components—such as occupancy scoring or local planning—limiting adaptability to unforeseen map structures or dynamic obstacles. A more flexible, DRL-driven frontier scoring and selection could better handle complex layouts.
	\item \textbf{High-speed exploration with partial observability}: Mobile robots often face limited onboard sensing and partial maps. Methods relying heavily on global costmaps or dense sensor fusion can stagnate if those costmaps are incomplete. Meanwhile, pure DRL policies can fail to generalize or exhibit slow learning when confronted with entirely unstructured spaces.
\end{itemize}

\subsection{Contributions and Novelty}
In this paper, we propose a novel architecture, termed \emph{FH-DRL} (Frontier Heuristic Deep Reinforcement Learning), that combines \emph{heuristic frontier selection} with DRL-based local navigation to address the aforementioned gaps. Our key contributions are as follows:
\begin{enumerate}
	\item \textbf{Adaptive frontier identification via an exponential-hyperbolic scoring function:} We design a distance score that blends exponential and hyperbolic terms, complemented by an occupancy-based stochastic measure to prioritize frontiers effectively. This addresses limitations in classical distance- or occupancy-only methods~\cite{Lubanco2020, Cimurs2022}.
	\item \textbf{Seamless integration of DRL into local navigation:} A Twin Delayed DDPG (TD3) controller operates continuously within the partially known map, offering robust obstacle avoidance while retaining high speeds. This mitigates slowdown issues commonly observed in standard global planners that rely on static costmaps.
	\item \textbf{Extensive validation in simulation and real environments:} We thoroughly evaluate \emph{FH-DRL} in ROS2 and Gazebo simulations, as well as real-world testing on a Turtlebot3 \texttt{waffle\_pi}. Comparisons against state-of-the-art frontier-based (NF, CFE) and DRL-based (GDAE) exploration methods show significant reductions in exploration time and travel distance across various complexity levels.
	\item \textbf{Scalable approach for intelligent vehicles and beyond:} Owing to the modular combination of heuristic frontier detection and DRL-based local planning, \emph{FH-DRL} is readily transferable to broader robotic systems, such as autonomous ground vehicles, industrial automation platforms, and planetary rovers, which operate in partially mapped or changing environments.
\end{enumerate}

\subsection{Paper Outline}
The remainder of this paper is organized as follows. Section~\ref{sec_related_work} reviews frontier-based exploration and DRL-based approaches, highlighting recent hybrid methods. Section~\ref{sec_fast_heuristics_algorithm} details our \emph{Frontier Heuristic Algorithm}, introducing the exponential-hyperbolic distance score and occupancy stochastic measure. Section~\ref{sec_fh_ANGUS} describes the \emph{FH-DRL} architecture, explaining how heuristic frontier selection and TD3-based control coalesce. In Section~\ref{sec_experiments}, we present simulation and real-world experiments, while Section~\ref{sec_conclusion} concludes with key insights, limitations, and avenues for future work.

\begin{figure*}[!hbp]
	\centering
	\includegraphics[width=0.75\linewidth]{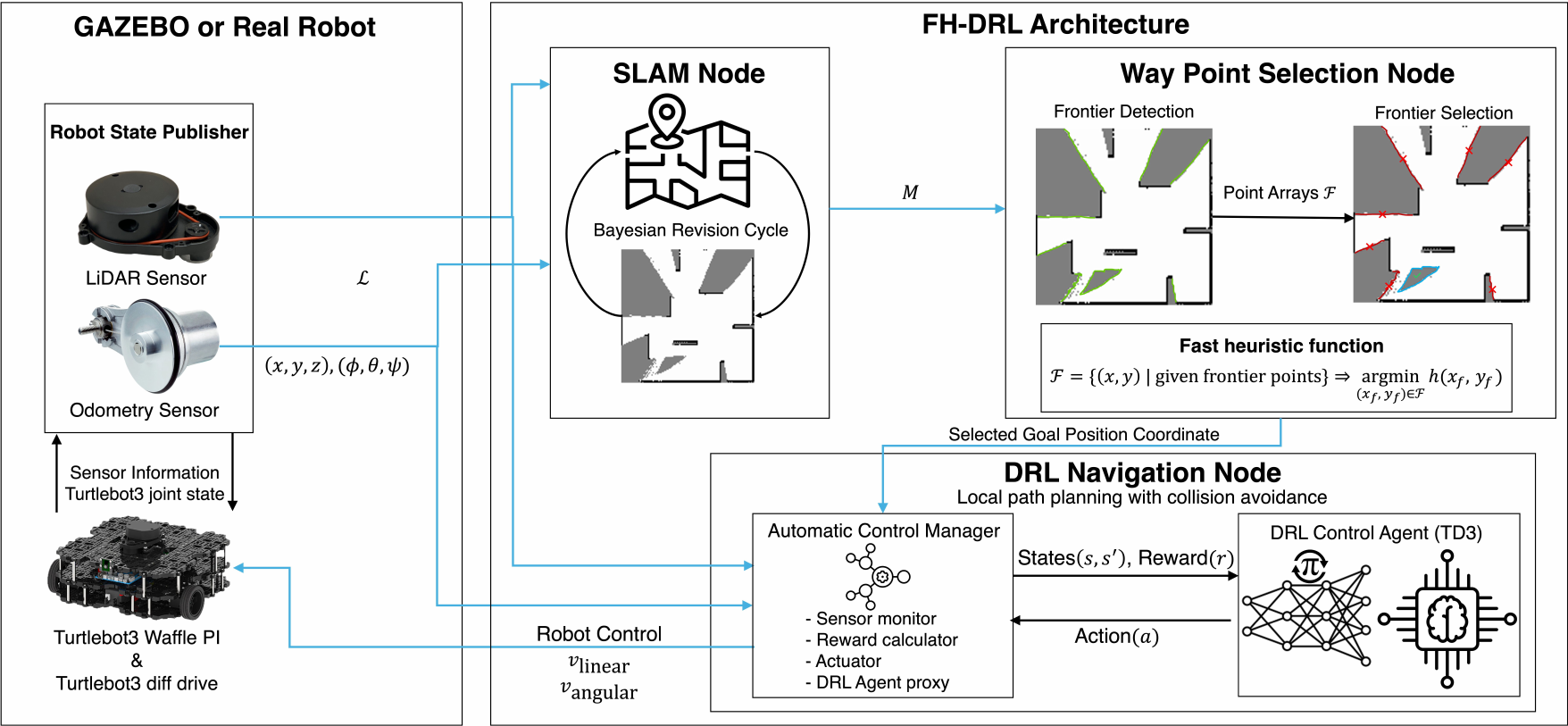}
	\caption{Overall Robot Navigation for Exploration Architecture}
	\label{fig_simulation_system_overall_architecture}
\end{figure*}

\section{Related Work}
\label{sec_related_work}

Research in autonomous exploration can be broadly grouped into \textbf{(A)} frontier-based approaches, \textbf{(B)} deep reinforcement learning (DRL)-based methods, and \textbf{(C)} hybrid solutions that integrate classical frontier-based heuristics with learning frameworks. Table~\ref{tab_comparison_related_work} compares representative works (including the proposed \emph{FH-DRL}) according to their primary focus, exploration strategy, key contributions, and evaluation settings.

\subsection{Frontier-Based Exploration}
Early pioneering work by Yamauchi~\cite{Yamauchi1997} introduced the notion of frontiers---boundaries between known and unknown regions---for iteratively driving the robot into uncharted areas. Subsequent extensions addressed inefficiencies in original frontier detection. For example, Holz \emph{et al.}~\cite{Holz2010} proposed clustering of frontier points to reduce computational overhead in large maps, while Basilico and Amigoni~\cite{Basilico2011} introduced multi-criteria decision making to balance map coverage against task objectives. Gao \emph{et al.}~\cite{Gao2018} and Lubanco \emph{et al.}~\cite{Lubanco2020} respectively refined frontier-selection heuristics by incorporating robot heading information and clustering for managing multiple frontiers in cluttered or open corridors.

Further improvements tackle incremental or dynamic frontier detection. Keidar and Kaminka~\cite{Keidar2012} proposed wavefront-based frontier detectors for faster updates, and Senarathne \emph{et al.}~\cite{SENARATHNE2015189} discussed incremental algorithms for safe and reachable frontier detection in real-time. Meanwhile, Liu \emph{et al.}~\cite{Liu2023} introduced heuristics-biased sampling to reduce both map coverage time and repeated visits. Despite these advances, purely frontier-driven methods often lack deeper adaptive capabilities when the search space or dynamic obstacles grow more complex~\cite{julia2012comparison}.

\subsection{DRL-Based Exploration}
Deep reinforcement learning enables robots to learn navigation and exploration policies directly from experience, mitigating reliance on prior map information or carefully hand-engineered heuristics. Niroui \emph{et al.}~\cite{Niroui2019} demonstrated end-to-end DRL exploration in cluttered rescue environments, showing faster coverage than classical approaches but also encountering slow convergence. Peake \emph{et al.}~\cite{Peake2021} employed DRL to adaptively switch between goal-directed and random exploration behaviors, boosting coverage under sensor noise. Cao \emph{et al.}~\cite{Cao2024} scaled DRL to large domains via hierarchical policies, whereas Li \emph{et al.}~\cite{Li2020} tackled partially observable settings by introducing subgoals to guide exploration in unknown zones.

However, fully DRL-based planners risk incurring high sample complexity and difficulty in ensuring guaranteed frontier coverage. Large or complex environments can cause partial observability issues that hamper policy learning~\cite{Mansfield2024}. Even with advanced reward shaping~\cite{Yan2023} or intrinsic motivation~\cite{Feng2023}, purely learning-driven systems can stall in local minima or repeatedly revisit certain areas unless carefully trained over extensive simulation runs~\cite{Li2022}.

\subsection{Hybrid Approaches: Frontier-Driven and Learning-Based}
To overcome the limitations of purely heuristic frontier algorithms and purely DRL exploration, recent works have combined the two paradigms. Cimurs \emph{et al.}~\cite{Cimurs2022} introduced a goal-driven exploration strategy in which frontiers were generated in a classical manner, but the selection among candidate frontiers was refined by a learned policy. In a similar vein, Wang \emph{et al.}~\cite{Wang2024} adopted DRL to rank frontier points, thereby improving map coverage and reducing travel distance over conventional frontier-based heuristhics. Some studies incorporate advanced heuristics or cost functions---for instance, Chen \emph{et al.}~\cite{Cheng2021} explored how heuristics can guide the DRL search space, accelerating the policy’s convergence and focusing the exploration on promising frontiers.

Despite these hybrid successes, open challenges remain. Many existing solutions rely on domain-specific heuristics or single-criterion cost functions, failing to balance the trade-offs among path length, coverage, and dynamic obstacle avoidance~\cite{Cimurs2021,Fan2023}. Moreover, fewer works provide \emph{full integration} of frontier detection, local DRL-based motion planning, and continuous path re-optimization. Therefore, there exists a gap in seamlessly unifying robust frontier selection methods with an adaptive local planner that can consistently navigate around obstacles at higher speeds.

\subsection{Our Contribution}
The proposed \emph{FH-DRL} architecture aligns with these hybrid approaches, yet it introduces several unique features. First, an \textit{exponential-hyperbolic distance score} couples short-range feasibility with long-range gain, a departure from conventional distance- or occupancy-only frontier heuristics. Second, it deploys DRL (via Twin Delayed DDPG) in the \emph{local navigation loop}, ensuring continuous obstacle avoidance without reliance on prior costmaps. Furthermore, \emph{FH-DRL} evaluates occupancy stochastics within a customizable circular region around each frontier, catering to real-time sensor constraints. As shown later, these design decisions collectively advance exploration efficiency, coverage rate, and motion smoothness compared to existing frontier-only or purely DRL approaches.

\begin{table*}[!htp]
	\centering
	\caption{Representative Works in Frontier-Based and Learning-Based Exploration, Contrasted with \emph{FH-DRL}.}
	\label{tab_comparison_related_work}
	\renewcommand{\arraystretch}{1.1}
	\resizebox{0.98\linewidth}{!}{
		\begin{tabular}{l l l l l c}
			\toprule
			\textbf{Reference} & \textbf{Year} & \textbf{Approach Type} & \textbf{Key Contributions} & \textbf{Evaluation} & \textbf{Focus}\\
			\midrule
			Yamauchi~\cite{Yamauchi1997} & 1997 & Frontier-based & Foundational frontier definition & Simulation + Real & Coverage\\
			Holz~\cite{Holz2010} & 2010 & Frontier-based & Frontier clustering for large maps & Simulation & Efficiency \\
			Basilico~\cite{Basilico2011} & 2011 & Frontier-based & Multi-criteria approach in rescue ops & Simulation & Multi-objective \\
			Lubanco~\cite{Lubanco2020} & 2020 & Frontier-based & Clustering + utility/cost for mobile robots & Simulation & Efficiency\\
			Niroui~\cite{Niroui2019} & 2019 & DRL-based & DRL in complex, cluttered rescue settings & Simulation & Unstructured environ. \\
			Peake~\cite{Peake2021} & 2021 & DRL-based & Adaptive exploration policy for UAV & Simulation & UAV-based \\
			Cimurs~\cite{Cimurs2022} & 2022 & Hybrid (Frontier + DRL) & Goal-driven frontier selection w/ DRL & Simulation & Large-scale coverage\\
			Wang~\cite{Wang2024} & 2024 & Hybrid (Frontier + DRL) & Deep RL ranks frontier goals & Simulation + Real & Shorter exploration paths\\
			Fan~\cite{Fan2023} & 2023 & Hybrid (Edge-based + RL) & Hierarchical path planner w/ frontier RL & Simulation & Outer-planet exploration\\
			Liu~\cite{Liu2023} & 2023 & Frontier-based & Heuristics-biased sampling for coverage & Simulation + Real & Efficiency \\
			Cao~\cite{Cao2024} & 2024 & DRL-based & Large-scale exploration w/ hierarchical RL & Simulation & Scalability\\
			\textbf{FH-DRL (Ours)} & \textbf{2024} & \textbf{Hybrid} & \begin{tabular}[c]{@{}l@{}}Exponential-hyperbolic frontier scoring; \\ DRL-based local navigation (TD3); \\ Occupancy stochastic function\end{tabular} & \begin{tabular}[c]{@{}l@{}}Simulation\\ + Real\end{tabular} & \begin{tabular}[c]{@{}l@{}}Improved coverage, \\ dynamic avoidance\end{tabular}\\
			\bottomrule
	\end{tabular}}
\end{table*}

\section{Frontier Heuristic Algorithm}
\label{sec_fast_heuristics_algorithm}

The Frontier Heuristic Algorithm, introduced in the Way-Point Selection Node (see Figure~\ref{fig_simulation_system_overall_architecture}), addresses the challenge of prioritising frontier points in unknown environments by synthesising spatial and occupancy-based criteria. Specifically, the algorithm dynamically combines a distance-based score, $\mathcal{D}(x_f, y_f)$, and an occupancy score, $\mathcal{O}(x_f, y_f)$, to quantify the exploration value of each frontier, thereby minimising redundant paths and unnecessary exploration.

The distance-based score, $\mathcal{D}(x_f, y_f)$, is optimised to reflect the constraints of both linear and angular robot velocities. It employs a customised “exponbolic” function that adjusts the score across three intervals—close-range, proportional, and far-range distances—thus ensuring balanced evaluation over multiple scales. Concurrently, the occupancy-based score, $\mathcal{O}(x_f, y_f)$, interprets local grid maps to determine the proportion of unknown and obstacle-occupied cells around each frontier, factoring in both environmental openness and occupancy density.

\begin{algorithm}
	\footnotesize
	\caption{Frontier Detection}
	\label{alg_frontier_detetion}
	\begin{algorithmic}[1]
		\STATE Initialise $\mathcal{F}$ as size of $M$ array, filled with \textbf{Non-Frontier} marks
		\STATE Set $w$ and $h$ as width and height of given map
		\FOR{$i = 0$ \textbf{to} $w - 1$ \textbf{and} $j = 0$ \textbf{to} $h - 1$}
		\IF{$M[i + w \cdot j]$ is $\varsigma_f$}
		\IF{$i > 0$ \textbf{and} $M[(i - 1) + w \cdot j]$ is $\varsigma_o$}
		\STATE $\mathcal{F}[i + w \cdot j] \gets f$
		\ELSIF{$i < w - 1$ \textbf{and} $M[(i + 1) + w \cdot j]$ is $\varsigma_u$}
		\STATE $\mathcal{F}[i + w \cdot j] \gets f$
		\ELSIF{$j > 0$ \textbf{and} $M[i + w \cdot (j - 1)]$ is $\varsigma_u$}
		\STATE $\mathcal{F}[i + w \cdot j] \gets f$
		\ELSIF{$j < h - 1$ \textbf{and} $M[i + w \cdot (j + 1)]$ is $\varsigma_u$}
		\STATE $\mathcal{F}[i + w \cdot j] \gets f$
		\ENDIF
		\ENDIF
		\ENDFOR
		\RETURN $\mathcal{F}$
	\end{algorithmic}
\end{algorithm}

Equation~\ref{eq_heuristic_function} presents the final heuristic function, $h(x_f, y_f)$, which combines both scores through a tunable discount factor $\gamma$. Larger values of $\mathcal{O}(x_f, y_f)$ primarily indicate open areas, whereas lower values reflect regions dominated by unknown spaces. Consequently, the algorithm favors frontiers with lower heuristic values, thereby selecting exploration objectives that effectively balance navigational efficiency with frontier exploration.

\subsection{Frontier Detection}
The frontier detection procedure, outlined in Algorithm~\ref{alg_frontier_detetion}, requires the occupancy grid map $M$ from the SLAM node. The map $M$ designates free space ($\varsigma_f$), obstacles ($\varsigma_o$), and unknown space ($\varsigma_u$). The detection algorithm searches for $\varsigma_u$ cells adjacent to $\varsigma_f$ or $\varsigma_o$, expands them in eight directions, and labels them as frontier points with a value $f$. Ultimately, the resulting set of frontier cells $\mathcal{F}$ is returned as an array at the conclusion of the frontier detection process.

\subsection{Exponential-Hyperbolic Distance Score}
A straightforward linear distance metric between the robot and a frontier is inadequate because it does not account for factors such as the unknown extent of the exploration environment or the robot’s limited linear and angular velocities. To address these constraints, we introduce the Exponential-Hyperbolic Distance Score, formulated in Equation~\ref{eq_exponential-hyperbolic_distance_score}.

\vspace*{-0.5cm}
\begin{equation}
    \text{\fontsize{8}{10}\selectfont$\mathcal{D}(x_f, y_f) = \tanh \Big(e^{\frac{d(x_f, y_f)}{\beta}} \cdot \sigma \big(e^{\frac{d(x_f, y_f)}{\beta}} \cdot (1 - \operatorname{csch} \frac{d(x_f, y_f)}{\alpha})\big)\Big)$}
\label{eq_exponential-hyperbolic_distance_score}
\end{equation}

Here, $\tanh$ and $\operatorname{csch}$ denote the hyperbolic tangent and hyperbolic cosecant functions, respectively, while $e$ is Euler’s number. The Euclidean distance is defined as $d(x_f, y_f) = \sqrt{(x_r - x_f)^2 + (y_r - y_f)^2}$, where $(x_r, y_r)$ represent the robot’s coordinates and $(x_f, y_f)$ corresponds to the centroid of the frontier. This expression interleaves exponential and hyperbolic terms, yielding a score dominated by the hyperbolic component at shorter distances and by the exponential component at longer distances. Additionally, the sigmoid normalisation $\sigma(x) = \frac{1}{1 + e^{-x}}$ confines the output to the interval $(0, 1)$.

The function partitions scoring into three regions:
\begin{itemize}
    \item A region that converges to 0, mitigating the effect of small distance disadvantages.
    \item A region in which the score grows with distance, balancing the trade-off between distance and $\mathcal{O}(x_f, y_f)$ (see Section~\ref{sec_occupancy_stochastic_score}).
    \item A region that approaches 1, effectively penalising distances beyond a specified threshold.
\end{itemize}

The parameters $\alpha$ and $\beta$ regulate the overall behavior of the function: $\alpha$ determines where the slope begins, whereas $\beta$ sets its gradient $\bigl(\tfrac{\partial}{\partial d}d(x_f, y_f)\bigr)$. These parameters can be chosen based on the robot’s LiDAR range and average speed. For example, a robot equipped with a more extensive LiDAR coverage—enabling larger map updates per Bayesian cycle—may require a higher $\alpha$ to deprioritise frontiers at closer ranges. Conversely, a faster robot may benefit from a larger $\beta$ to capture its enhanced capacity for reaching distant frontiers.

\begin{figure}
	\centering
	\includegraphics[width=0.7\linewidth]{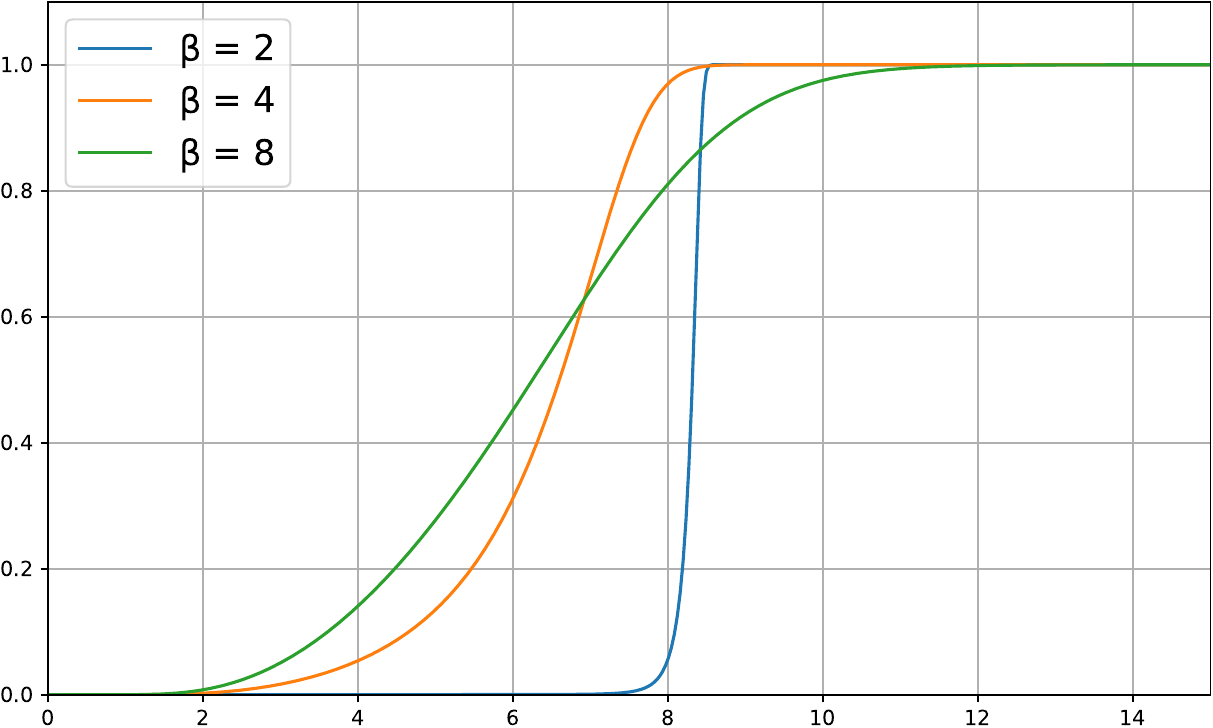}
	\caption{Exponential-Hyperbolic Distance Score}
	\label{fig_exponential_hyperbolic_distance_score}
\end{figure}

\begin{figure}
	\centering
	\begin{subfigure}[b]{0.45\linewidth}
		\centering
		\includegraphics[width=1.0\linewidth]{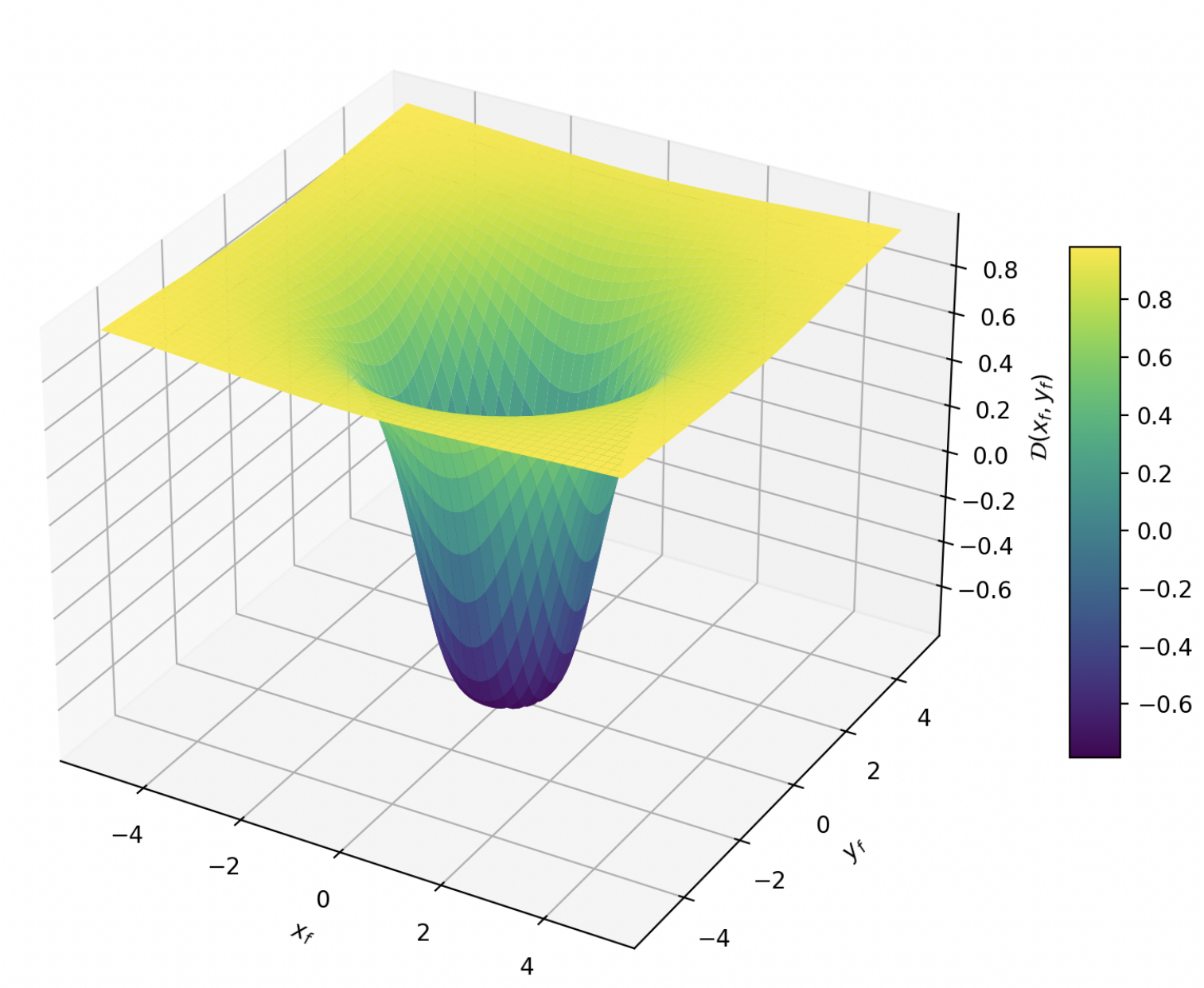}
		\caption{Exponential-Hyperbolic Distance Score}
		\label{fig_distance_score_3d}
	\end{subfigure}
	\hfill
	\begin{subfigure}[b]{0.53\linewidth}
		\centering
		\includegraphics[width=0.95\linewidth]{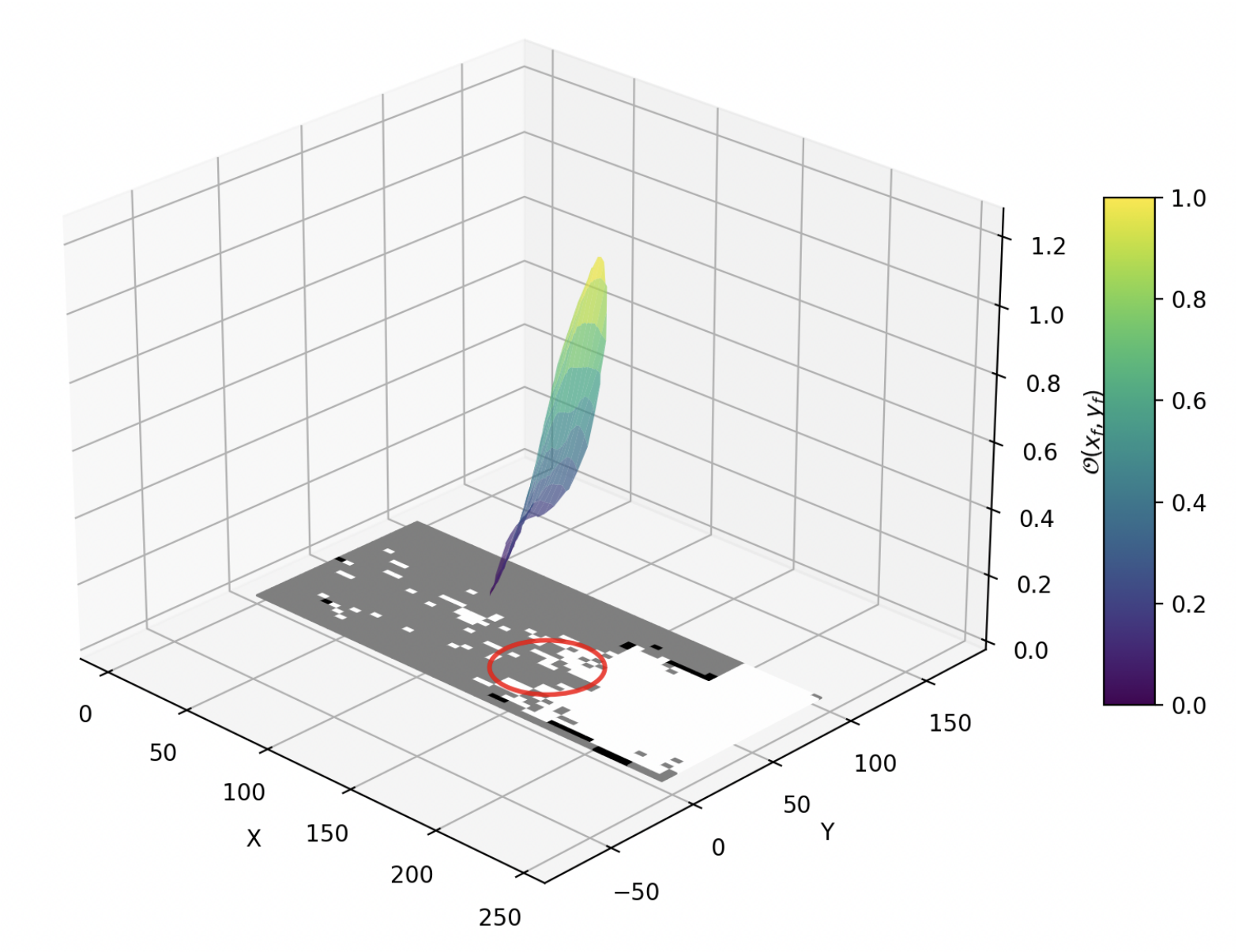}
		\caption{Occupancy Grid Score}
		\label{fig_occupancy_gird_score_3d}
	\end{subfigure}
	\caption{Visualisation of exponential-hyperbolic distance and occupancy grid score distribusion}
	\label{fig_representative_frontiers}
\end{figure}

The equation’s raw form can be difficult to deploy directly because the slope pivots near the $\tfrac{1}{2}$ mark, complicating how $\alpha$ and $\beta$ are interpreted. To address this, an additional exponential term stabilises the function, setting the pivot at the maximal score of 1. Lastly, the $\tanh$ normalisation ensures a smooth, bounded output. The resultant Exponential-Hyperbolic Distance Score is illustrated in Figure~\ref{fig_exponential_hyperbolic_distance_score}, with a 3D visualisation of the score’s spatial distribution relative to the robot’s location in Figure~\ref{fig_distance_score_3d}.

\subsection{Occupancy Stochastic Score}
\label{sec_occupancy_stochastic_score}

The Occupancy Stochastic Score, as the name implies, evaluates the occupancy grid map in the vicinity of a frontier to determine its exploratory value. The score is computed according to Equation~\ref{eq_occupancy_stochastic_score}:

\vspace*{-0.3cm}
\begin{align}
	& \mathcal{O}\bigl(x_f, y_f\bigr)= \frac{\sum_{(x, y) \in S} m_{(x, y)}}{\pi r^2} \cdot \operatorname{sech}(a_f), \label{eq_occupancy_stochastic_score} \\
	& S = \Bigl\{(x, y)\;\Big|\;(x - x_f)^2 + (y - y_f)^2 \le r^2 \Bigr\}, \label{eq_occupancy_stochastic_score_set}
\end{align}

where $\operatorname{sech}$ is the hyperbolic secant function, $\bigl(x_f, y_f\bigr)$ designates the centroid of the frontier, and $a_f$ is the frontier length. $m_{(x, y)}$ is the occupancy value at the position $(x, y)$, defined over the set $S$ in Equation~\ref{eq_occupancy_stochastic_score_set}. Unlike GDAE, which employs a discretised square kernel, the FH-DRL architecture computes continuous obstacle likelihood within a circular region, thereby enhancing stability through orientation-sensitive and smoothly probabilistic modelling that includes unknown, free, and occupied cells in the frontier.

The function $m(x, y)$ represents the occupancy value at a given point $(x, y)$ on the map. Within the \texttt{costmap2D} infrastructure of ROS2, inflation propagates costs outward from occupied cells with diminishing intensity proportional to the distance. We define five distinct categories of cost relevant to the robot’s movement:

\begin{itemize}
	\item \texttt{"Lethal"} cost indicates an actual obstacle cell with a high probability of collision at the robot’s centre.
	\item \texttt{"Inscribed"} cost denotes a cell lying inside the robot’s inscribed radius from an obstacle, ensuring collision if the robot’s centre occupies it.
	\item \texttt{"Possibly circumscribed"} cost implies that collision may occur based on the robot’s orientation.
	\item \texttt{"Freespace"} cost is zero, indicating no known obstacles.
	\item \texttt{"Unknown"} cost conveys no available information about the cell, leaving its interpretation open.
\end{itemize}

Other costs lie between \texttt{"Freespace"} and \texttt{"Possibly circumscribed"} as a function of their distance from a \texttt{"Lethal"} cell and a user-defined decay function. This flexible scheme allows planners to consider the robot’s footprint only when orientation is critical. Each cell in the map has up to 255 probability values. By remapping 255 (unknown) to 0 and incrementing all other values by 1, unknown cells are set to 0 while those with higher obstacle likelihood obtain larger values. Cells confirmed to contain obstacles (probability 1) take on the maximum value of 255, signifying a minimal exploratory benefit. We set the radius $r$ from the frontier centroid to its outermost frontier cell to assess the exploration value. Summing the probability values for all cells inside this circle, then dividing by the circle’s area, yields an average probability in the interval $[0,1]$. A value near 0 signifies predominantly unknown or free areas, whereas a value approaching 1 indicates mostly obstacle-occupied regions, which in turn lowers the exploration value. Finally, normalising these localised probability characteristics via $\operatorname{sech}(a_f)$ and multiplying by the frontier length yield $\mathcal{O}(x_f, y_f)$, which accounts for both openness and size. While GDAE employs a fixed kernel size for evaluation—leading to potential generalisation limitations—the FH-DRL framework dynamically adjusts the inscribed circular radius, facilitating more accurate heuristic estimation. A 3D depiction of the resulting $\mathcal{O}(x_f, y_f)$ distribution appears in Figure~\ref{fig_occupancy_gird_score_3d}.

\begin{figure*}[!htbp]
	\centering
	\begin{subfigure}[b]{0.265\linewidth}
		\centering
		\includegraphics[width=0.75\linewidth]{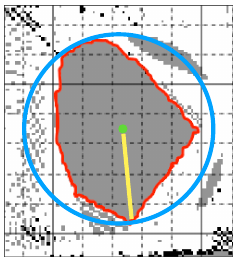}
		\caption{}
		\label{fig_closed_frontier}
	\end{subfigure}
	\hfill
	\begin{subfigure}[b]{0.265\linewidth}
		\centering
		\includegraphics[width=0.80\linewidth]{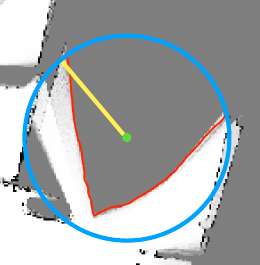}
		\caption{}
		\label{fig_open_wided_frontier_01}
	\end{subfigure}
	\hfill
	\begin{subfigure}[b]{0.225\linewidth}
		\centering
		\includegraphics[width=1.0\linewidth]{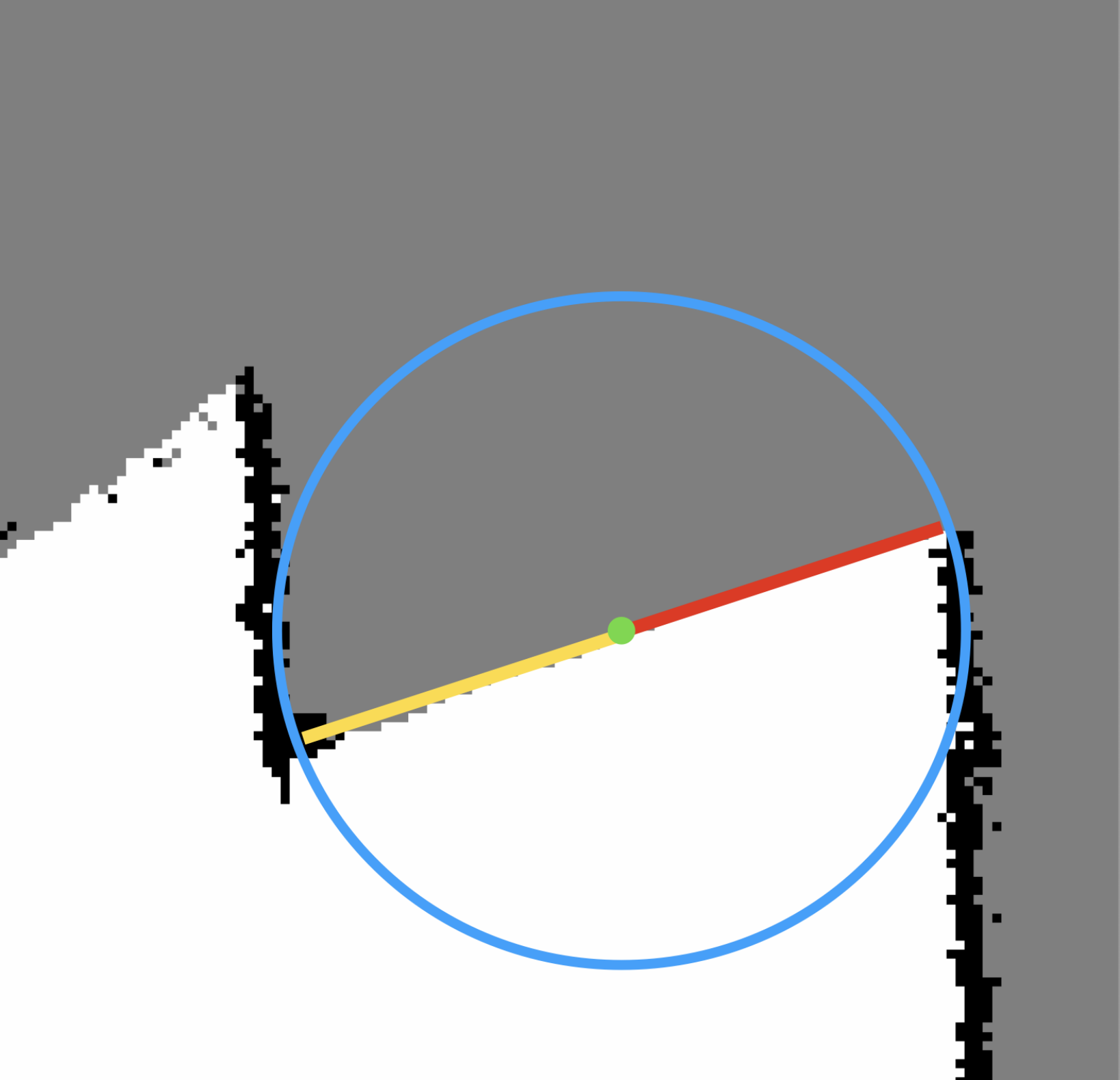}
		\caption{}
		\label{fig_door_gap_frontier}
	\end{subfigure}
	\hfill
	\begin{subfigure}[b]{0.18\linewidth}
		\centering
		\includegraphics[width=0.95\linewidth]{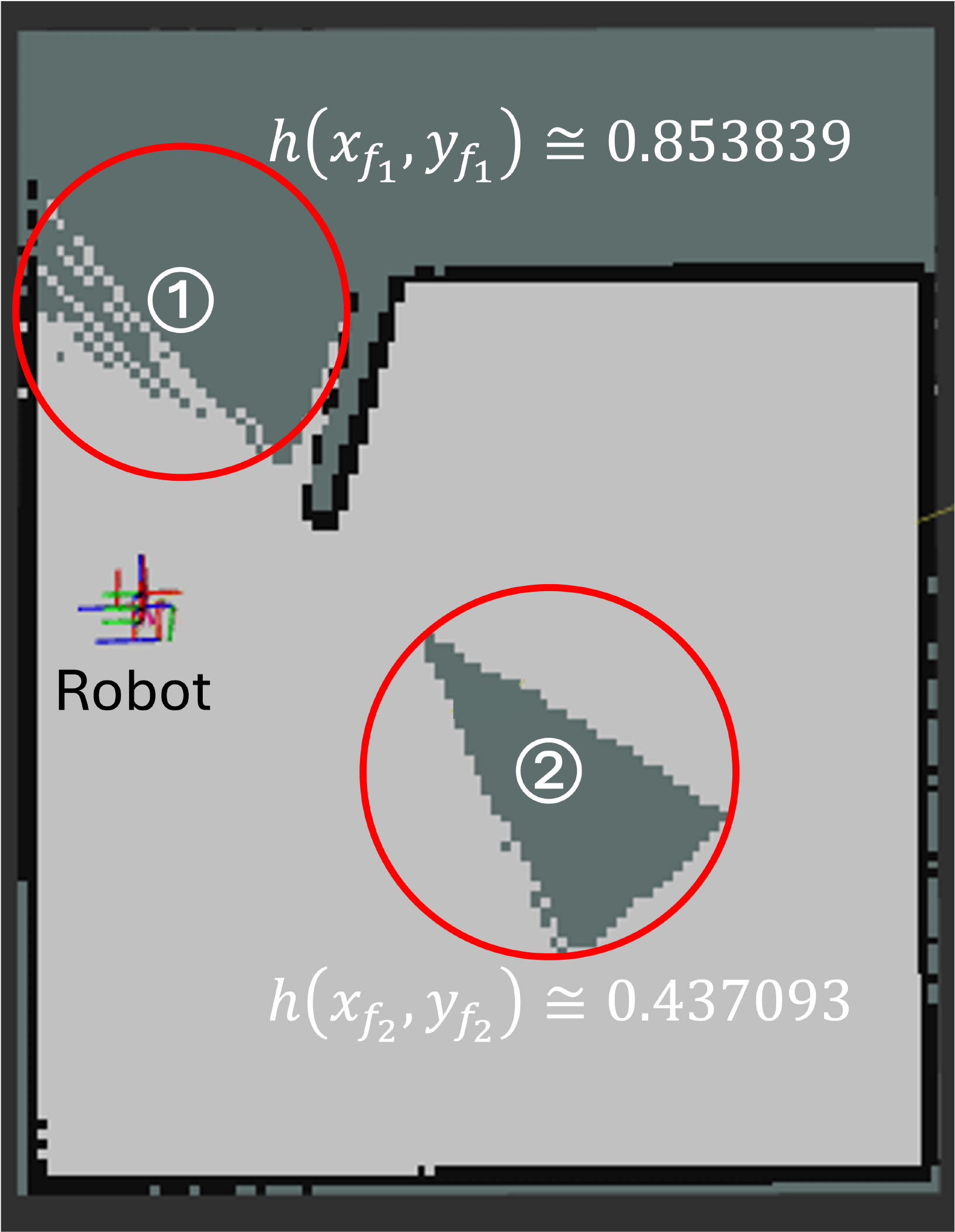}
		\caption{}
		\label{fig_example_scenario}
	\end{subfigure}
	\caption{Representative Frontiers \\ \footnotesize (a). Closed Frontier, (b). Open Wided Frontiers, (c). Door Gap Frontier, (d) Case-study}
	\label{fig_representative_frontiers}
\end{figure*}

\subsection{Frontier Types}

Frontier points in the exploration space can be classified into three main categories according to their spatial configuration.

\paragraph{Closed Frontier}
As depicted in Figure~\ref{fig_closed_frontier}, closed frontiers generally emerge between regions that have already been explored, forming gaps that remain uncharted. Giving these frontiers a higher priority enhances exploration efficiency by obviating the need for subsequent returns. Their defining elements include a red boundary denoting the frontier line, a green centroid, a yellow marker indicating the farthest point from the centroid, and a blue circle whose radius matches the distance to that yellow point. Because unknown space predominantly occupies the interior of this circle, the resulting $\mathcal{O}(x_f, y_f)$ value tends toward 0, making it the lowest among the three frontier types.

\paragraph{Open Wide Frontier}
Open wide frontiers occur in expansive areas perceived by the robot’s LiDAR sensor, extending beyond its immediate range threshold, as illustrated in Figure~\ref{fig_open_wided_frontier_01}. Prioritising these frontiers often enhances exploration efficiency, since they typically align with the robot’s primary direction of travel. Owing to a mixture of unknown cells, free space, and occasional obstacles, $\mathcal{O}(x_f, y_f)$ is higher here than in closed frontiers, rendering them the second priority.

\paragraph{Door Gap Frontier}
As shown in Figure~\ref{fig_door_gap_frontier}, door gap frontiers arise where obstacles—such as opened doors—create narrow passageways between explored areas. These frontiers commonly appear during transitions between spaces or when altering navigation paths. To promote efficiency, this type should be assigned the lowest priority if the present area remains only partially explored. Due to their relatively higher ratio of free space interspersed with some unknown zones, $\mathcal{O}(x_f, y_f)$ reaches its maximum among the three frontier types, making it the last choice for exploration.

\begin{figure}
	\centering
	\includegraphics[width=0.7\linewidth]{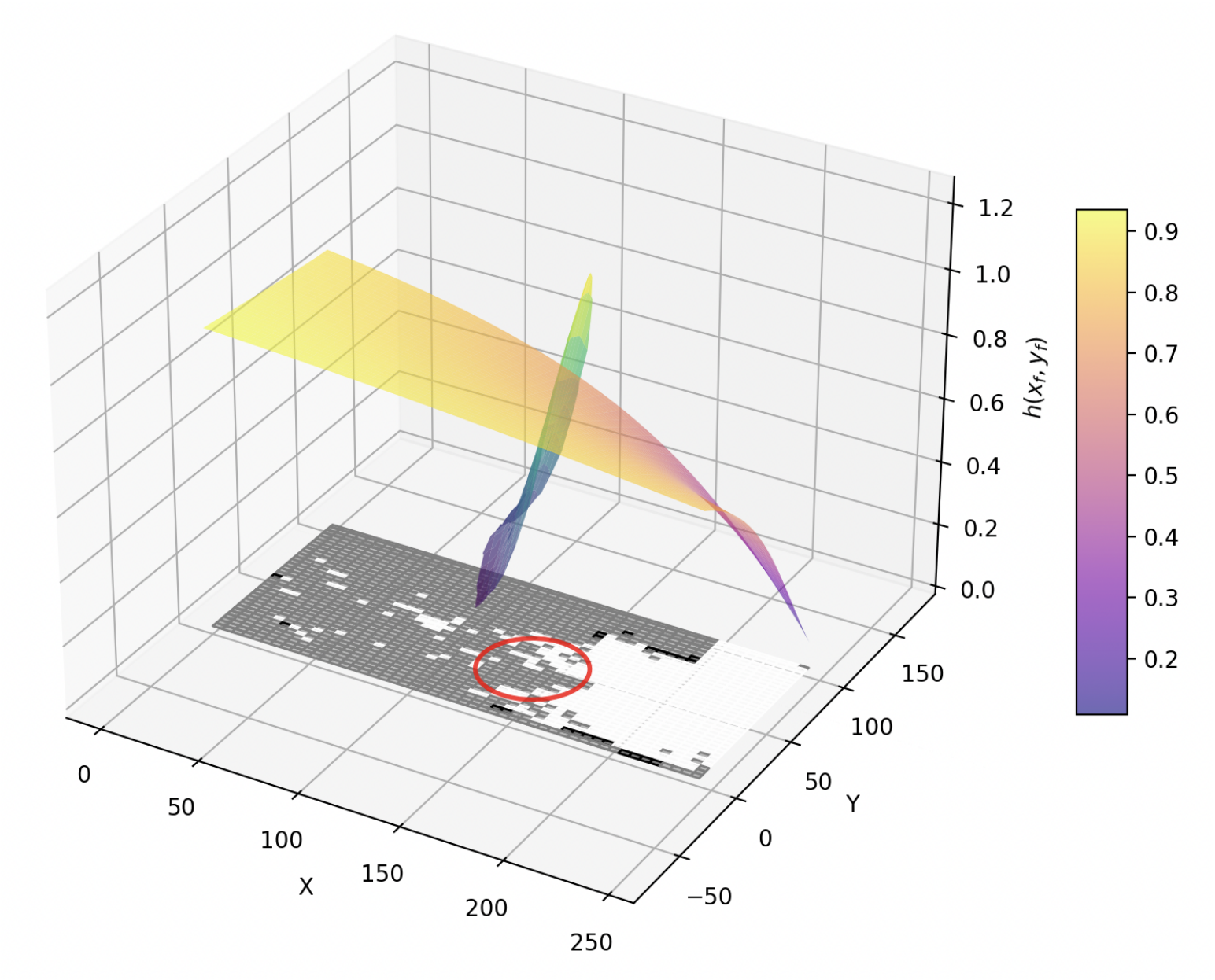}
	\caption{FH-DRL score 3D Distribution}
	\label{fig_fh-angus_score}
\end{figure}

\subsection{Frontier Heuristic Function}

A fast heuristic function harnesses empirical knowledge to reduce both search paths and computation time, thereby expediting and enhancing the process of finding optimal solutions. At close distances, the corresponding score converges to 0, ensuring negligible distinctions among nearby points. Beyond a certain threshold, the derivative of this function rises steeply before decreasing. Consequently, $\mathcal{D}(x_f, y_f)$ can be subdivided into three intervals: close-range, proportional range, and far-range. By integrating the scores from $\mathcal{D}(x_f, y_f)$ and $\mathcal{O}(x_f, y_f)$, one obtains the final heuristic function defined in Equation~\ref{eq_heuristic_function}, whose three-dimensional distribution is depicted in Figure~\ref{fig_fh-angus_score}.


\vspace*{-0.5cm}
\begin{equation}
		h(x_f, y_f) \equiv \mathcal{D}(x_f, y_f) \cdot \gamma + \mathcal{O}(x_f, y_f) \cdot (1 - \gamma)
	\label{eq_heuristic_function}
\end{equation}

In this formulation, the parameter $\gamma$ serves as a discount factor that modulates the heuristic function under the constraint $\gamma \le \tfrac{1}{2}$. By combining distance-based and occupancy-based scores, the heuristic function determines the overall heuristic value of a given frontier point. Specifically, $\mathcal{D}(x_f, y_f)$ is scaled by $\gamma$, and $\mathcal{O}(x_f, y_f)$ is scaled by $1 - \gamma$. The frontier goal is then selected by minimizing this heuristic score, i.e., \texttt{Frontier Goal} $= \argmin h(f_i)$, where $f_i$ denotes the set of candidate frontier points $(x_f, y_f)$ that integrate both distance and occupancy information for prioritizing valuable frontiers.

As illustrated in Figure~\ref{fig_example_scenario}, consider two frontiers, Frontier~1 and Frontier~2. Frontier~1 is located between an open area and a door gap that may require further exploration. Conventional methods such as Nearest Frontier (NF), which focus primarily on proximity, would initially explore Frontier~1 and subsequently return to Frontier~2. However, under FH-DRL, Frontier~2 is prioritized due to its lower heuristic score. Its enclosed geometry, characterized by a higher ratio of unknown space and an extended frontier length, results in a reduced score. Consequently, FH-DRL avoids redundant paths by favoring Frontier~2.

\section{FH-DRL Architecture}
\label{sec_fh_ANGUS}

\subsection{Overall Architecture}
The FH-DRL (Frontier Heuristic Deep Reinforcement Learning) architecture, depicted in Figure~\ref{fig_simulation_system_overall_architecture}, is designed to facilitate autonomous navigation in a GAZEBO simulation by integrating several pivotal components. The Robot State Publisher provides essential data derived from a LiDAR sensor scanning $n(\mathcal{L}) = 360$ at $2\pi$ radians, along with odometry tracking (\texttt{Pose:} $x, y, z$; $\phi, \theta, \psi$) and joint states. The SLAM node concurrently updates the map via a Bayesian Revision Cycle. The Way-Point Selection node identifies unexplored regions, generates arrays of candidate points, and applies a heuristic function ($\mathcal{F} = h(x_f, y_f)$) to define local goals from the SLAM-generated map. Local path planning and collision avoidance fall under the DRL Navigation Node, which employs a TD3-based control agent for decision-making grounded in states ($s, s'$), rewards ($r$), and actions ($a$). The Automatic Control Manager processes sensor data for reward calculation and issues linear and angular velocity commands to the robot, facilitating both obstacle avoidance and goal-directed maneuvers. This architecture effectively integrates SLAM, frontier detection and selection, and deep reinforcement learning to achieve efficient autonomous navigation.


\subsection{Way-Point Selection Node}
The Way-Point Selection node determines feasible exploration targets by examining frontiers within the map $\mathcal{M}$, as described in Section~\ref{sec_fast_heuristics_algorithm}. Once the SLAM node provides $\mathcal{M}$, the algorithm extracts the frontiers, which define the boundaries between known and unknown regions. By integrating distance-based and occupancy-based scores, the heuristic function prioritises unexplored areas for each frontier. Specifically, the frontier score is computed from Equation~\ref{eq_heuristic_function}, which balances the distance component $\mathcal{D}(x_f, y_f)$ against the occupancy score $\mathcal{O}(x_f, y_f)$ according to the heuristic parameters. The frontier exhibiting the lowest heuristic value is then chosen as the robot’s next waypoint, thereby optimising the trade-off between exploration speed and safety.

\subsection{SLAM Node}

Simultaneous Localisation and Mapping (SLAM) is essential for autonomous robot navigation while enabling real-time map building and localisation. In this study, the SLAM node uses the \texttt{nav2} package in ROS2 to construct an occupancy grid which representing navigable areas and obstacles based on LiDAR sensor data.

SLAM iteratively refines the map and the robot's position by combining sensor data with motion estimates through a Bayesian revision cycle. The Adaptive Monte Carlo Localisation (AMCL) algorithm utilises particle filters to further enhances positional accuracy. The \texttt{nav2} package integrates exploration and localisation using algorithms such as gmapping ensures up-to-date map generation for precise navigation in dynamic environments.

The SLAM node communicates the information to the Frontier Selection node to enabe the selection of optimal exploration points for continued navigation and efficient environment coverage as soon as the updated map is generated.

\subsection{DRL Navigation Node}
Deep reinforcement learning (DRL) facilitates local navigation that adapts continuously to unexpected obstacles, circumventing the limitations of conventional methods such as \texttt{nav2} in ROS2—which generally rely on pre-established maps. Through active interaction with the environment, DRL methods learn navigation policies that improve adaptability, obstacle avoidance, and maximum traversal speed, even in the absence of a map. This flexibility makes DRL a compelling alternative for enhancing autonomous navigation capabilities.

The Twin Delayed DDPG (TD3) algorithm, developed by Fujimoto \textit{et al.} \cite{fujimoto2018}, refines Deep Deterministic Policy Gradient (DDPG) by mitigating value overestimation in continuous action spaces, rendering it well-suited for self-navigation scenarios. Key innovations in TD3 include Clipped Double Q-learning, Delayed Policy Updates, and Target Policy Smoothing, all of which bolster its performance.

The Automatic Control Manager receives sensor data from GAZEBO and augments it with the robot’s previous actions. It then computes the reward and supplies both current and subsequent states to the TD3 agent. The actor uses the state vector $(s_t)$ to determine two-dimensional continuous actions, $\bigl(A_\text{linear}, A_\text{angular}\bigr)$. Meanwhile, during training, the critics receive the same input as the actor for their first fully connected layer, then incorporate the current-step actions in the second layer to generate Q-values.

\section{Experiments}
\label{sec_experiments}

\subsection{Simulation Setup}

\subsubsection{Scenario Settings}
\begin{figure*}[!htbp]
	\centering
	\begin{subfigure}[b]{0.17\linewidth}
		\centering
		\includegraphics[width=1.0\linewidth]{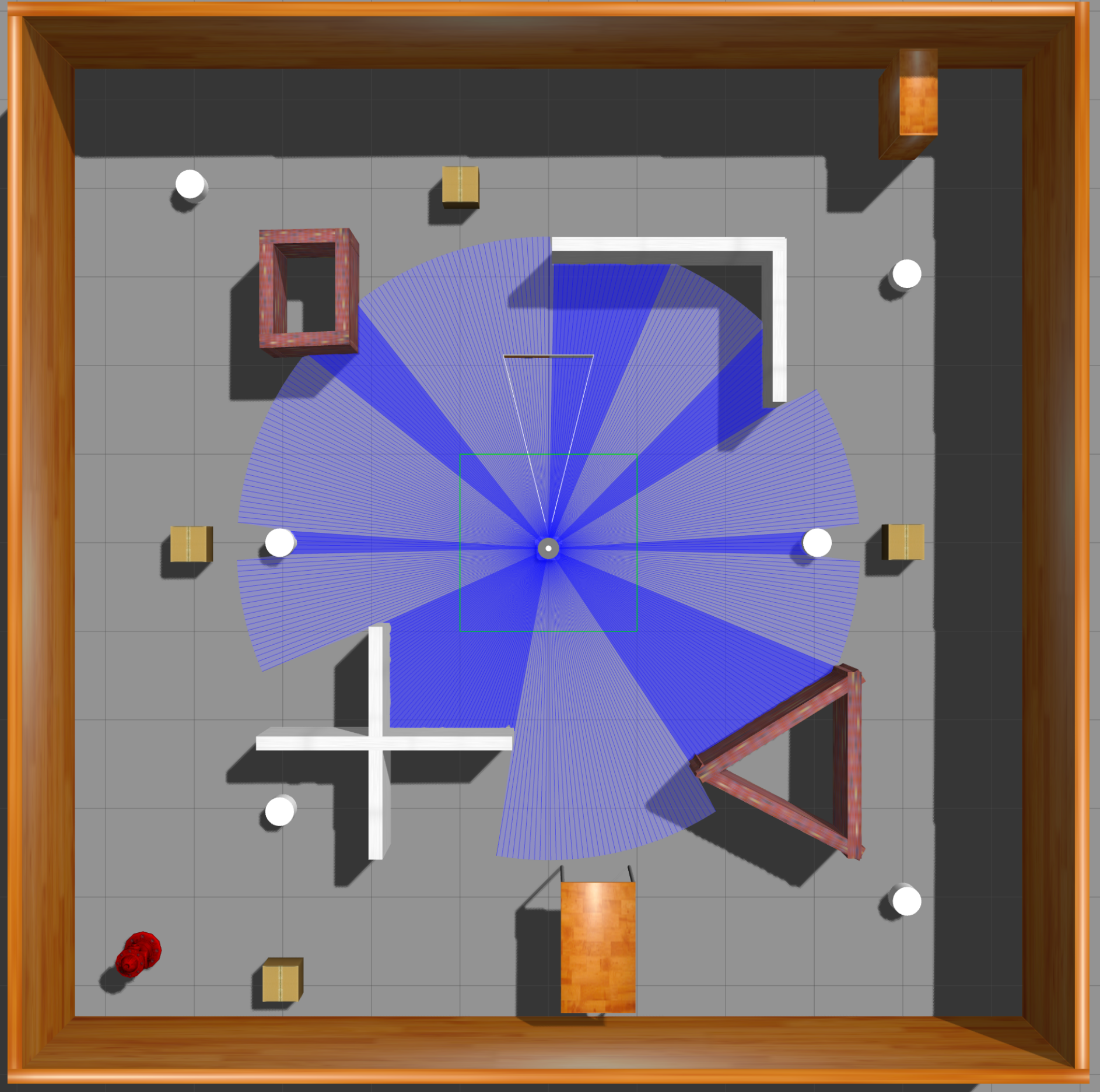}
		\caption{}
		\label{fig_training_environment}
	\end{subfigure}
	\hfill
	\begin{subfigure}[b]{0.17\linewidth}
		\centering
		\includegraphics[width=1.0\linewidth]{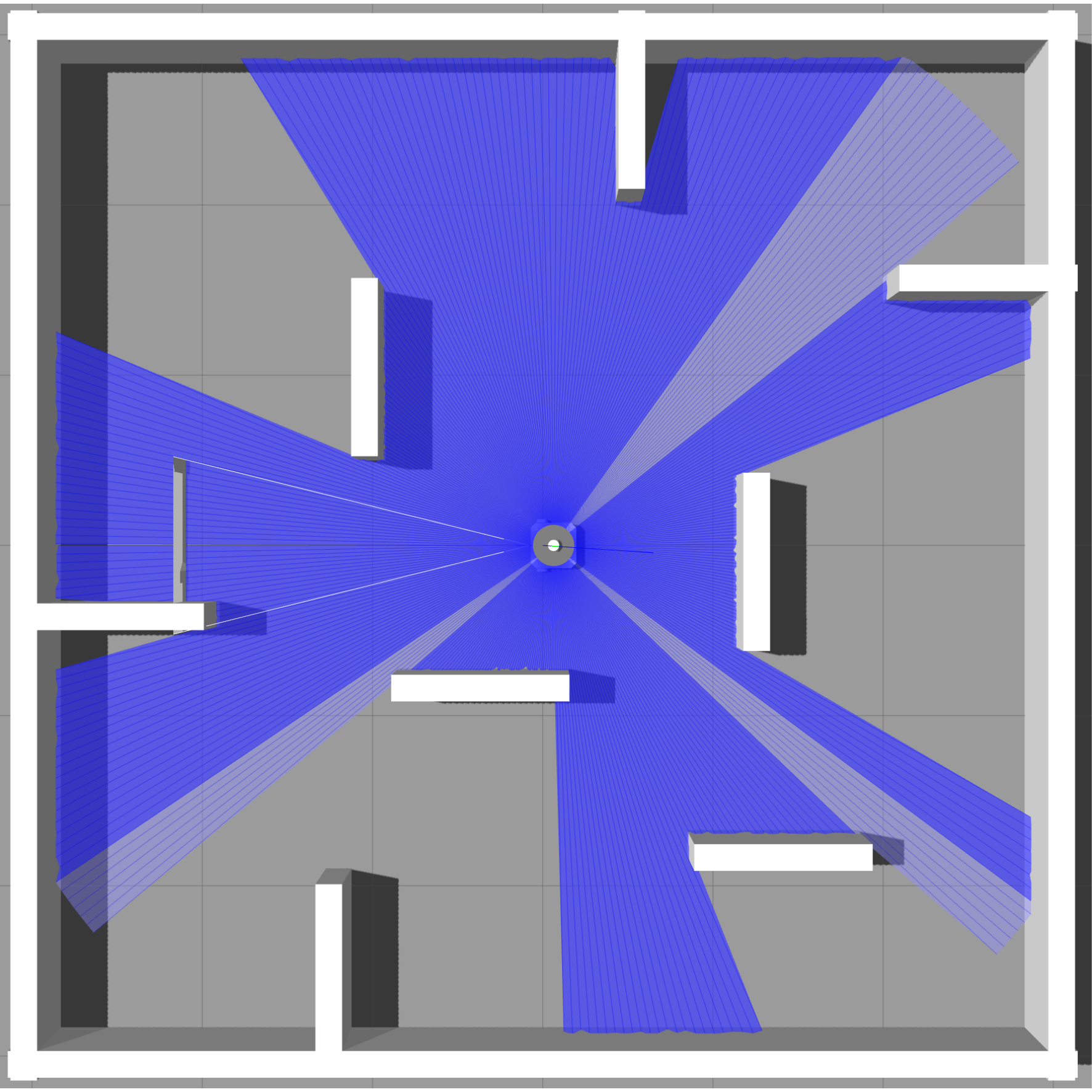}
		\caption{}
		\label{fig_simulation_environment_world_01}
	\end{subfigure}
	\hfill
	\begin{subfigure}[b]{0.3\linewidth}
		\centering
		\includegraphics[width=\linewidth]{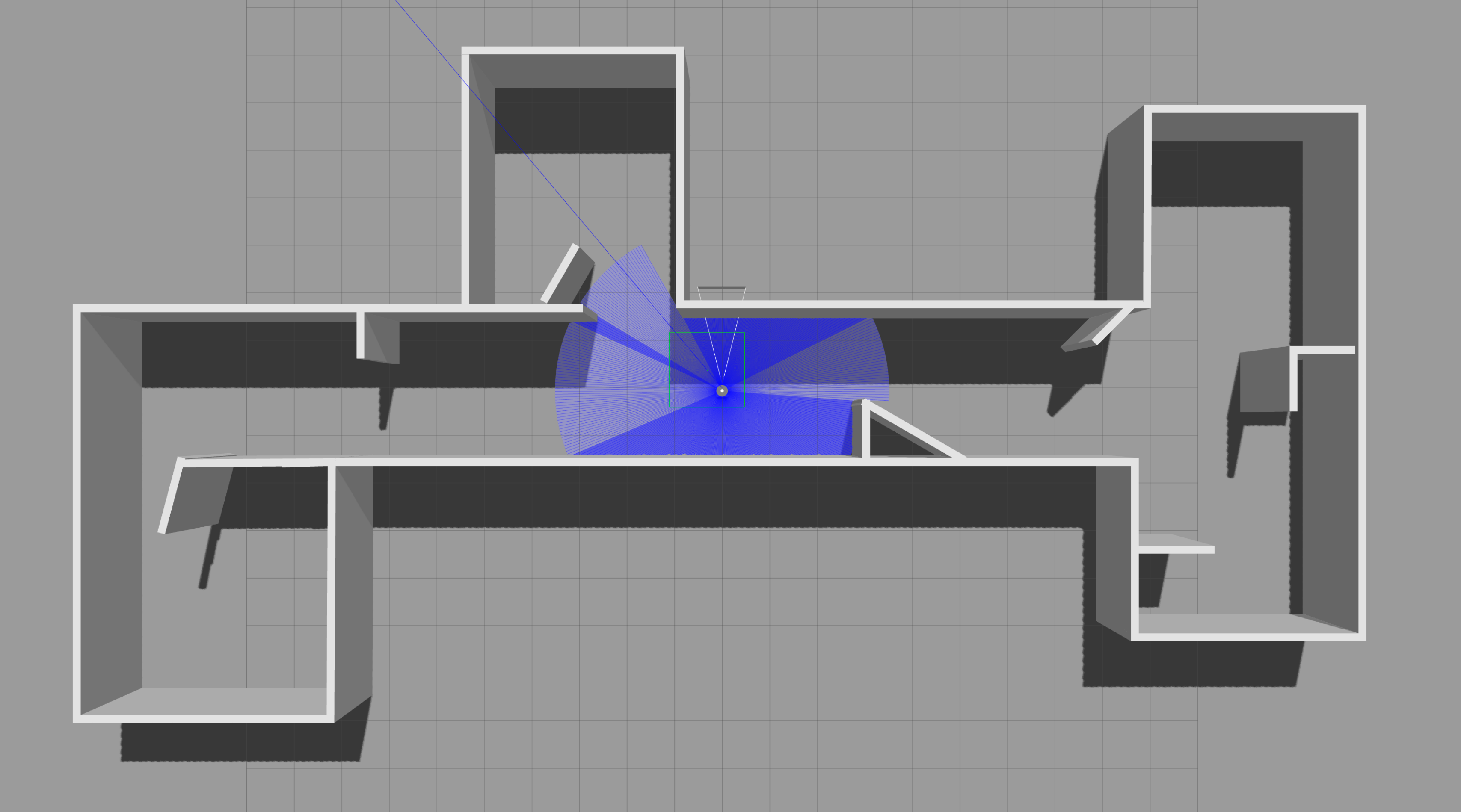}
		\caption{}
		\label{fig_simulation_environment_world_02}
	\end{subfigure}
	\hfill
	\begin{subfigure}[b]{0.305\linewidth}
		\centering
		\includegraphics[width=\linewidth]{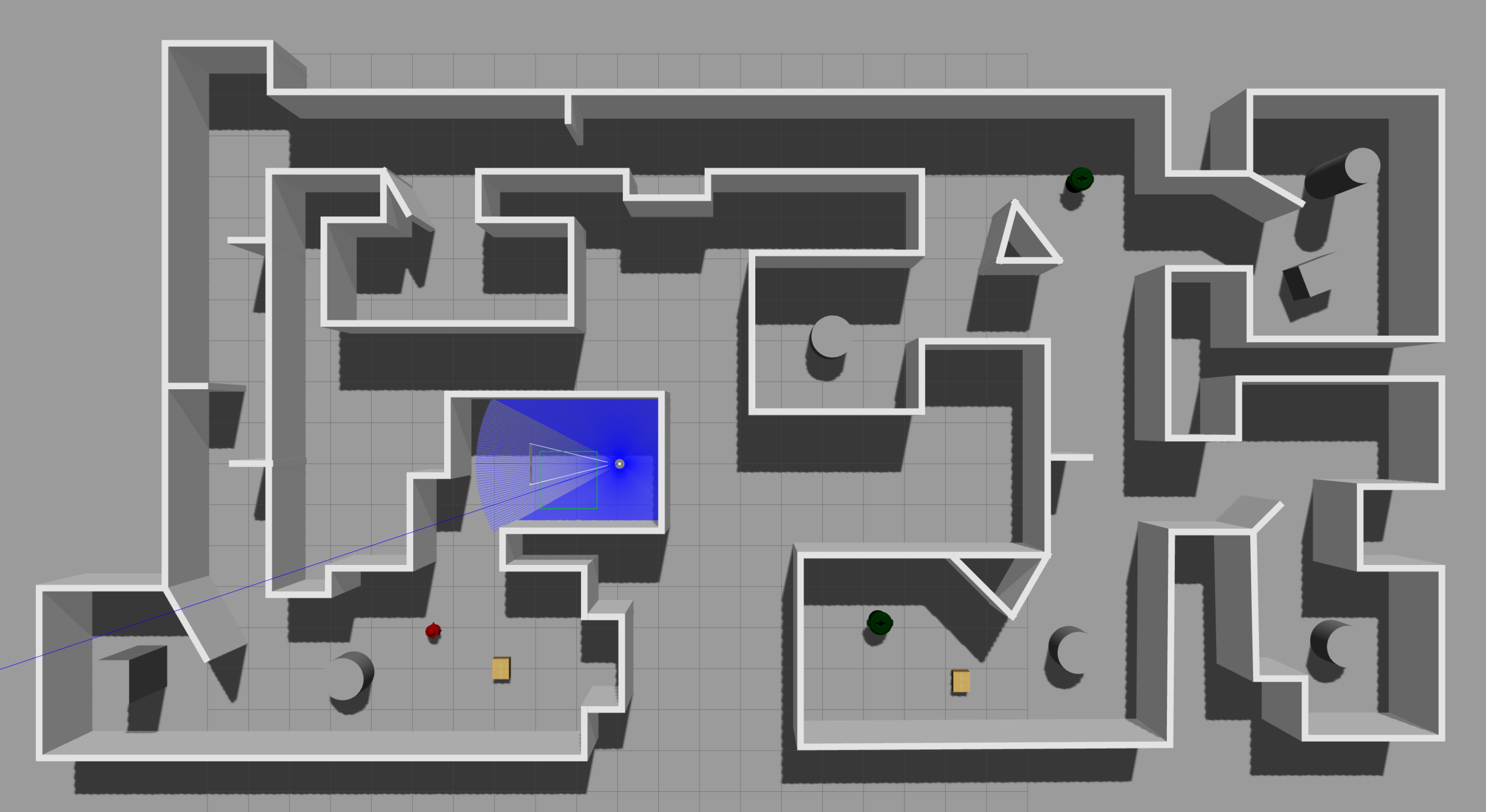}
		\caption{}
		\label{fig_simulation_environment_world_03}
	\end{subfigure}
	\caption{Simulation Scenarios. \footnotesize (a) Training Environment; (b) Small Space with Obstacles; (b) Corridor with Turns; (c) Complex Maze}
	\label{fig_simulation_environment_worlds}
\end{figure*}

We conducted comparative evaluations in a ROS2 and GAZEBO \texttt{Classic} environment using a Turtlebot3 \texttt{waffle\_pi}. The experimental scenarios are as follows. In Scenario~I (Low Complexity Map, see Figure~\ref{fig_simulation_environment_world_01}), the robot navigates a small enclosed area with internal walls. Scenario~II (Medium Complexity Map, see Figure~\ref{fig_simulation_environment_world_02}) tests its ability to traverse corridors containing turns and junctions, representative of indoor office-type settings. Scenario~III (High Complexity Map, see Figure~\ref{fig_simulation_environment_world_03}) introduces a maze-like layout with narrow passages, challenging advanced navigation and decision-making algorithms.

We compared frontier selection in FH-DRL against CFE \cite{Lubanco2020}, GDAE \cite{Cimurs2022}, and NF \cite{Yamauchi1997}. For local navigation, FH-DRL uses ROS2’s \texttt{nav2}. In order to isolate and compare only the frontier-selection capabilities, we replaced the original local path-planning component in GDAE (which was designed to reach a global goal in unknown space) with \texttt{nav2}, omitting the global goal from its heuristic function. Moreover, we evaluated FH-DRL both with and without the DRL navigation node: FH-DRL without DRL uses \texttt{nav2} for local control, whereas FH-DRL with DRL employs the proposed TD3-based controller. This setup enables an assessment of the impact of FH-DRL’s Frontier Heuristic (introduced in Section~\ref{sec_fast_heuristics_algorithm}) independent of any benefits from DRL-based navigation.

\begin{table}
	\centering
	\caption{Hyperparameter Values}
	\label{tab_Hyperparameter}
	\begin{tabular}{lr}
	\toprule
	\textbf{Parameter}              & \textbf{Value}  \\
	\midrule
	Neurons in actor network        & 3394            \\
	Neurons in critic network       & 1026            \\
	Batch size                     	& 512             \\
	Buffer size                    	& 1000000         \\
	Discount factor                	& 0.99            \\
	Learning rate                  	& 0.002           \\
	Tau                             & 0.003           \\
	Step time                      	& 0.01            \\
	Epsilon decay               	& 0.9995          \\
	Epsilon minimum                	& 0.05            \\
	Policy noise                  	& 0.2             \\
	Policy noise clip             	& 0.5             \\
	Policy update frequency       	& 2               \\
	\bottomrule
	\end{tabular}
\end{table}

\subsubsection{Neural Networks and Hyperparameter Settings}
The actor network depicted in Table \ref{tab_Hyperparameter} is implemented as a fully connected neural network (FCNN) with six hidden layers, whereas the critic network consists of three hidden layers. The actor processes normalized LiDAR sensor inputs, denoted by $n(\mathcal{L}) = 360$ and rescaled to the interval $[0,1]$, in addition to state inputs such as the initial goal distance $d_{g_\text{init}}$, goal orientation $\angle g_\text{angle}$, and previous linear and angular actions, $A_\text{linear}^{t-1}$ and $A_\text{angular}^{t-1}$. The critic uses these same state inputs, with its second fully connected layer also incorporating the actor’s actions ($A_\text{linear}$ and $A_\text{angular}$) to compute $Q$-values. Based on these estimated $Q$-values, the actor ultimately outputs the optimal linear and angular actions. Table~\ref{tab_Hyperparameter} details the hyperparameters employed during training.

\subsubsection{Model Training}
\begin{figure}[h!]
    \centering
    \begin{subfigure}[b]{0.47\linewidth}
        \centering
        \includegraphics[width=\linewidth]{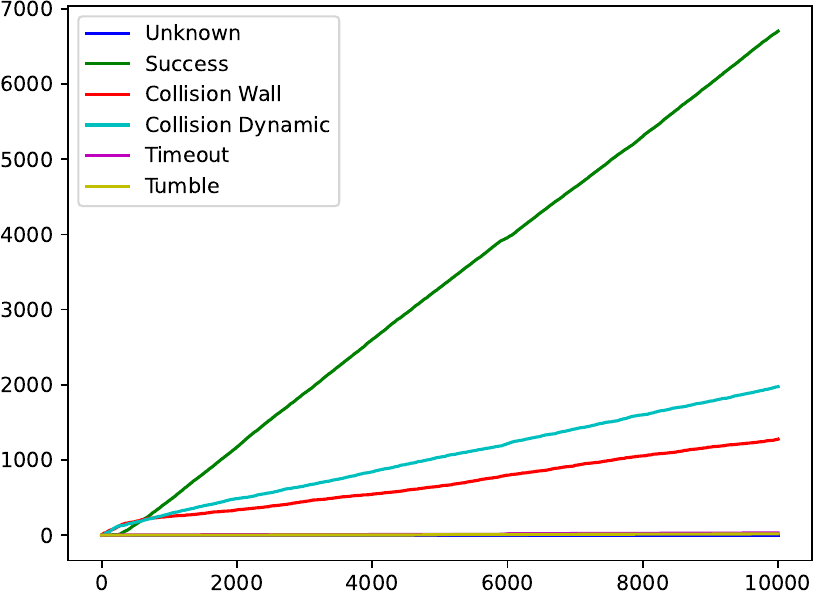}
        \caption{Outcomes}
        \label{fig_outcomes}
    \end{subfigure}
    \hfill
    \begin{subfigure}[b]{0.47\linewidth}
        \centering
        \includegraphics[width=\linewidth]{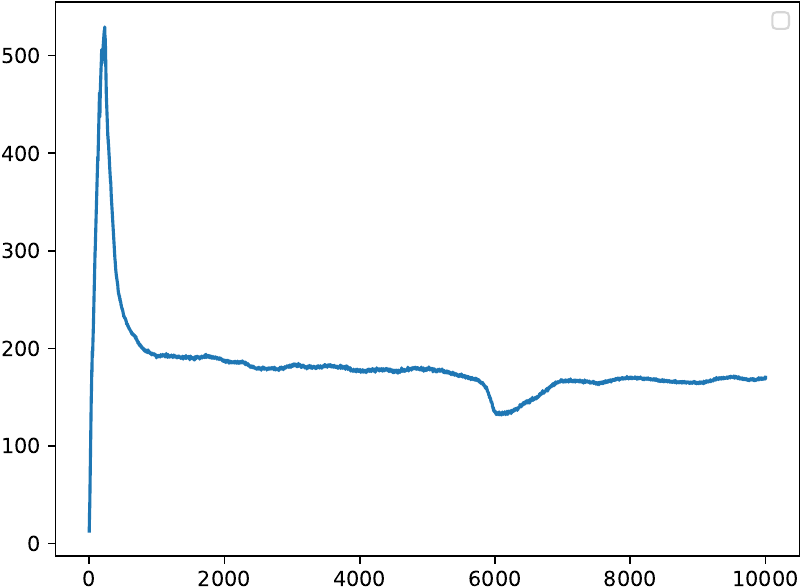}
        \caption{Actor Loss}
        \label{fig_actor_loss}
    \end{subfigure}

    \vskip\baselineskip
    \begin{subfigure}[b]{0.47\linewidth}
        \centering
        \includegraphics[width=\linewidth]{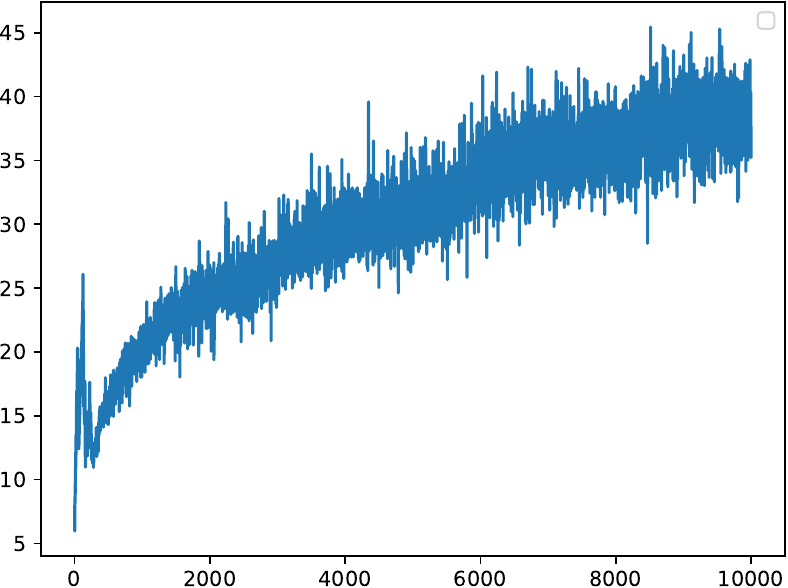}
        \caption{Critic Loss}
        \label{fig_critic_loss}
    \end{subfigure}
    \hfill
    \begin{subfigure}[b]{0.47\linewidth}
        \centering
        \includegraphics[width=\linewidth]{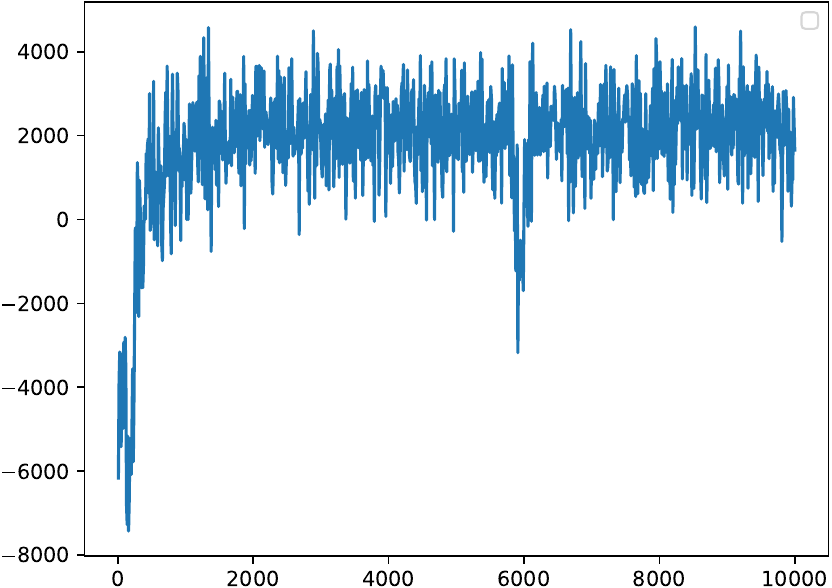}
        \caption{Average Reward}
        \label{fig_average_reward}
    \end{subfigure}

    \caption{Training Progress}
    \label{fig_training_progress}
\end{figure}

The TD3 algorithm for local navigation was implemented in a GAZEBO simulation featuring six dynamic obstacles. The training spanned 10,000 episodes, conducted on a system equipped with an \texttt{Intel i5-10400F CPU} and an \texttt{NVIDIA 3070 GPU}, and required approximately two days to complete. Goals were randomly generated through the GAZEBO goal generator and forwarded to the DRL navigation node. As illustrated in Figure~\ref{fig_training_environment}, the training environment incorporates both static and dynamic obstacles, simulating complex navigation tasks. Training progress is visualized in Figure~\ref{fig_training_progress}.

At the start of each episode, the robot is initialized either at the central area of the environment or at the previous goal location. Upon receiving a randomly generated goal, the robot proceeds to navigate while avoiding obstacles. An episode concludes unsuccessfully if the robot collides with an obstacle or exceeds its allotted time, triggering a penalty. By contrast, successful goal completion results in a positive reward, in accordance with Algorithm~\ref{alg_reward_function_DRL}, thereby incentivizing faster goal attainment.

The TD3 reward function presented in Algorithm~\ref{alg_reward_function_DRL} integrates various criteria to guide exploration in unknown environments, rewarding both obstacle avoidance and efficient goal-reaching behavior. Positive rewards are issued if the Euclidean distance to the goal ($d_{g_\text{robot}}$) is less than the initial distance ($d_{g_\text{init}}$) or is within the threshold $T_g$. Penalties arise if the robot fails to maintain its maximum linear velocity ($M_\text{linear}$), minimum angular action ($A_\text{angular}$), or a suitable orientation ($\angle g_\text{angle}$). Additional penalties occur whenever $\min_{l \in \mathcal{L}} l$—where $l \in \mathcal{L}$—falls below the collision threshold ($T_c$) or below $1.5 \cdot T_c$ if the robot is dangerously close to a potential collision zone.

\begin{algorithm}
	\footnotesize
	\caption{Reward Function of DRL Algorithm}
	\label{alg_reward_function_DRL}
	\begin{algorithmic}[1]
		\STATE $r_{\text{yaw}} \gets -|\angle g_\text{angle}|$ 
		\STATE $r_{\text{linear}} \gets -((M_\text{linear} - A_\text{linear}) \cdot 10)^2$
		\STATE $r_{\text{angular}} \gets -A_\text{angular}^2$ 
		\STATE $r_{\text{distance}} \gets \frac {2 \cdot d_{g\_\text{init}}} {d_{g\_\text{init}} + d_{g\_\text{robot}} - 1}$ 
		\IF{$\min\limits_{l \in \mathcal{L}} l < 1.5 \cdot T_c$} 
		\STATE $r_\text{obstacle} \gets -50$
		\ELSE
		\STATE $r_\text{obstacle} \gets 0$
		\ENDIF
		\STATE $R \gets \sum \{r_{\text{yaw}},\ r_{\text{angular}},\ r_{\text{distance}},\ r_\text{obstacle}\}$
		\IF{$d_{g\_\text{robot}} < T_g$}
		\STATE $R \gets R + 5000$
		\ENDIF
		\IF{$\min\limits_{l \in \mathcal{L}} l < T_c$}
		\STATE $R \gets R - 2000$
		\ENDIF
		\RETURN $R$
	\end{algorithmic}
\end{algorithm}

\subsection{Experimental Simulation Results and Analyses}

\subsubsection{Environment Reveal}
Table~\ref{tab_Merged_Experiment_Results} compares the performance of multiple autonomous exploration methods across different experimental scenarios. The primary metrics encompass average, maximum, minimum, and standard deviation ($\sigma$) values for distance (d, in meters), time (T, in seconds), and exploration rate (ExpR, in percent). Each scenario was evaluated by running 10 independent trials.

\paragraph{Scenario I (Low Complexity Map):}
FH-DRL without DRL achieves the most favorable results, with an average distance of 13.15\,m and a completion time of 71.2\,s. Moreover, it exhibits notably low variance in travel distance (0.67\,m). FH-DRL with DRL performs comparably well (16.66\,m, 81.28\,s) and attains a high exploration rate (98.03\%). While Nearest Frontier (NF) and Cognet Frontier Exploration (CFE) also yield high exploration rates (98.37\% and 98.69\%, respectively), they require significantly greater distances and times—15.47\,m and 135.27\,s for NF, and 22.16\,m and 123.06\,s for CFE. By contrast, GDAE exhibits the longest average time (153.47\,s) and a relatively high mean distance (16.96\,m), despite maintaining a comparable exploration rate (98.25\%).

\paragraph{Scenario II (Medium Complexity Map):}
FH-DRL with DRL outperforms the alternatives by achieving the shortest average completion time (325.09\,s) and a high exploration rate (99.47\%). Although its average distance (76.75\,m) slightly exceeds that of FH-DRL without DRL (69.48\,m), the latter also offers strong performance (352.70\,s, 99.24\%). In contrast, NF and CFE both require substantially longer times: 542.69\,s and 545.52\,s, respectively. GDAE exhibits a high degree of variability, posting an average time of 880.59\,s and a distance of 78.61\,m.

\paragraph{Scenario III (High Complexity Map):}
FH-DRL with DRL again emerges as the leading approach, posting the shortest average time (785.34\,s) and the highest exploration rate (99.74\%). Its average distance (175.77\,m) remains well below those recorded by NF and CFE. Meanwhile, FH-DRL without DRL also achieves solid results (889.19\,s, 99.56\%). In comparison, NF (2610.87\,s, 210.73\,m) and CFE (1552.86\,s, 268.33\,m) demand notably higher times and distances. Although GDAE covers a shorter distance (211.40\,m), it records an exceptionally long average time (1251.90\,s). Taken together, these findings confirm that FH-DRL, both with and without DRL, substantially outperforms the other methods—particularly in environments of greater complexity.

\begin{table*}[!ht]
	\centering
	\caption{Experimental Results in Completing Environment Exploration}
	\label{tab_Merged_Experiment_Results}
	\scriptsize
	\setlength{\tabcolsep}{3pt}
	\begin{tabular}{ccccccccccccccccccc}
		\toprule
		\multirow{2}{*}{Metrics} & \multicolumn{3}{c}{NF} & \multicolumn{3}{c}{CFE} & \multicolumn{3}{c}{GDAE} & \multicolumn{3}{c}{FH-DRL w/o DRL} & \multicolumn{3}{c}{FH-DRL w/ DRL} \\
		\cmidrule(lr){2-4} \cmidrule(lr){5-7} \cmidrule(lr){8-10} \cmidrule(lr){11-13} \cmidrule(lr){14-16}
		& I & II & III & I & II & III & I & II & III & I & II & III & I & II & III \\
		\midrule
		Avg dist.(m) & 15.471 & 83.897 & 210.731 & 22.162 & 91.692 & 268.330 & 16.958 & 78.611 & 211.397 & \textcolor{blue}{13.151} & \textcolor{blue}{69.480} & \textcolor{blue}{163.879} & \textcolor{red}{16.655} & \textcolor{red}{76.745} & \textcolor{red}{175.768} \\
		Min dist.(m) & 13.027 & 71.441 & 180.311 & 19.357 & 80.325 & 210.376 & 13.806 & 68.933 & 187.847 & \textcolor{blue}{12.170} & \textcolor{blue}{61.677} & \textcolor{blue}{153.073} & \textcolor{red}{16.655} & \textcolor{red}{67.926} & \textcolor{red}{161.479} \\
		Max dist.(m) & 17.540 & 95.460 & 244.956 & 25.998 & 104.449 & 337.185 & 20.388 & 91.310 & 267.180 & \textcolor{blue}{13.985} & \textcolor{blue}{80.995} & \textcolor{blue}{188.797} & \textcolor{red}{20.531} & \textcolor{red}{87.555} & \textcolor{red}{194.186} \\
		$\sigma$(dist) & 1.840 & 8.057 & 21.329 & 2.418 & 7.630 & 31.981 & 2.039 & 8.161 & 23.593 & \textcolor{blue}{0.667} & \textcolor{blue}{5.280} & \textcolor{blue}{10.566} & \textcolor{red}{2.717} & \textcolor{red}{7.150} & \textcolor{red}{12.324} \\
		Avg T.(sec) & 135.270 & 542.690 & 2610.870 & 123.060 & 545.520 & 1552.860 & 153.470 & 880.590 & 1251.900 & \textcolor{blue}{71.200} & \textcolor{blue}{352.700} & \textcolor{blue}{889.190} & \textcolor{red}{81.280} & \textcolor{red}{325.090} & \textcolor{red}{785.340} \\
		Min T.(sec) & 111.900 & 487.100 & 1115.100 & 109.400 & 490.200 & 1277.800 & 107.900 & 354.800 & 1051.600 & \textcolor{blue}{58.600} & \textcolor{blue}{310.100} & \textcolor{blue}{804.100} & \textcolor{red}{66.100} & \textcolor{red}{276.700} & \textcolor{red}{683.100} \\
		Max T.(sec) & 151.400 & 670.200 & 14167.100 & 140.500 & 588.100 & 1764.900 & 253.600 & 4545.000 & 1600.300 & \textcolor{blue}{86.100} & \textcolor{blue}{409.200} & \textcolor{blue}{1019.400} & \textcolor{red}{99.700} & \textcolor{red}{382.200} & \textcolor{red}{870.000} \\
		$\sigma$(T) & 13.887 & 53.853 & 4063.987 & 12.147 & 37.931 & 167.475 & 39.445 & 1290.434 & 163.315 & \textcolor{blue}{10.469} & \textcolor{blue}{30.809} & \textcolor{blue}{77.790} & \textcolor{red}{11.924} & \textcolor{red}{37.995} & \textcolor{red}{63.175} \\
		Avg ExpR.(\%) & 98.370 & 99.485 & 99.099 & 98.688 & 99.423 & 98.981 & 98.247 & 99.495 & 98.971 & \textcolor{blue}{97.784} & \textcolor{blue}{99.240} & \textcolor{blue}{99.557} & \textcolor{red}{98.025} & \textcolor{red}{99.466} & \textcolor{red}{99.735} \\
		Min ExpR.(\%) & 97.501 & 99.165 & 95.272 & 98.261 & 99.230 & 95.249 & 98.010 & 99.112 & 97.116 & \textcolor{blue}{97.486} & \textcolor{blue}{97.478} & \textcolor{blue}{99.236} & \textcolor{red}{96.453} & \textcolor{red}{99.042} & \textcolor{red}{99.539} \\
		Max ExpR.(\%) & 99.134 & 99.866 & 99.894 & 99.947 & 99.697 & 99.971 & 98.709 & 99.738 & 99.972 & \textcolor{blue}{98.041} & \textcolor{blue}{99.996} & \textcolor{blue}{99.956} & \textcolor{red}{99.719} & \textcolor{red}{99.880} & \textcolor{red}{99.988} \\
		$\sigma$(ExpR) & 0.502 & 0.244 & 1.463 & 0.594 & 0.134 & 1.944 & 0.176 & 0.189 & 1.128 & \textcolor{blue}{0.002} & \textcolor{blue}{0.670} & \textcolor{blue}{0.190} & \textcolor{red}{0.882} & \textcolor{red}{0.260} & \textcolor{red}{0.123} \\
		\bottomrule
	\end{tabular}
\end{table*}

\begin{figure*}[!htbp]
	\centering
	\begin{subfigure}[b]{0.3\linewidth}
		\centering
		\includegraphics[width=1.0\linewidth]{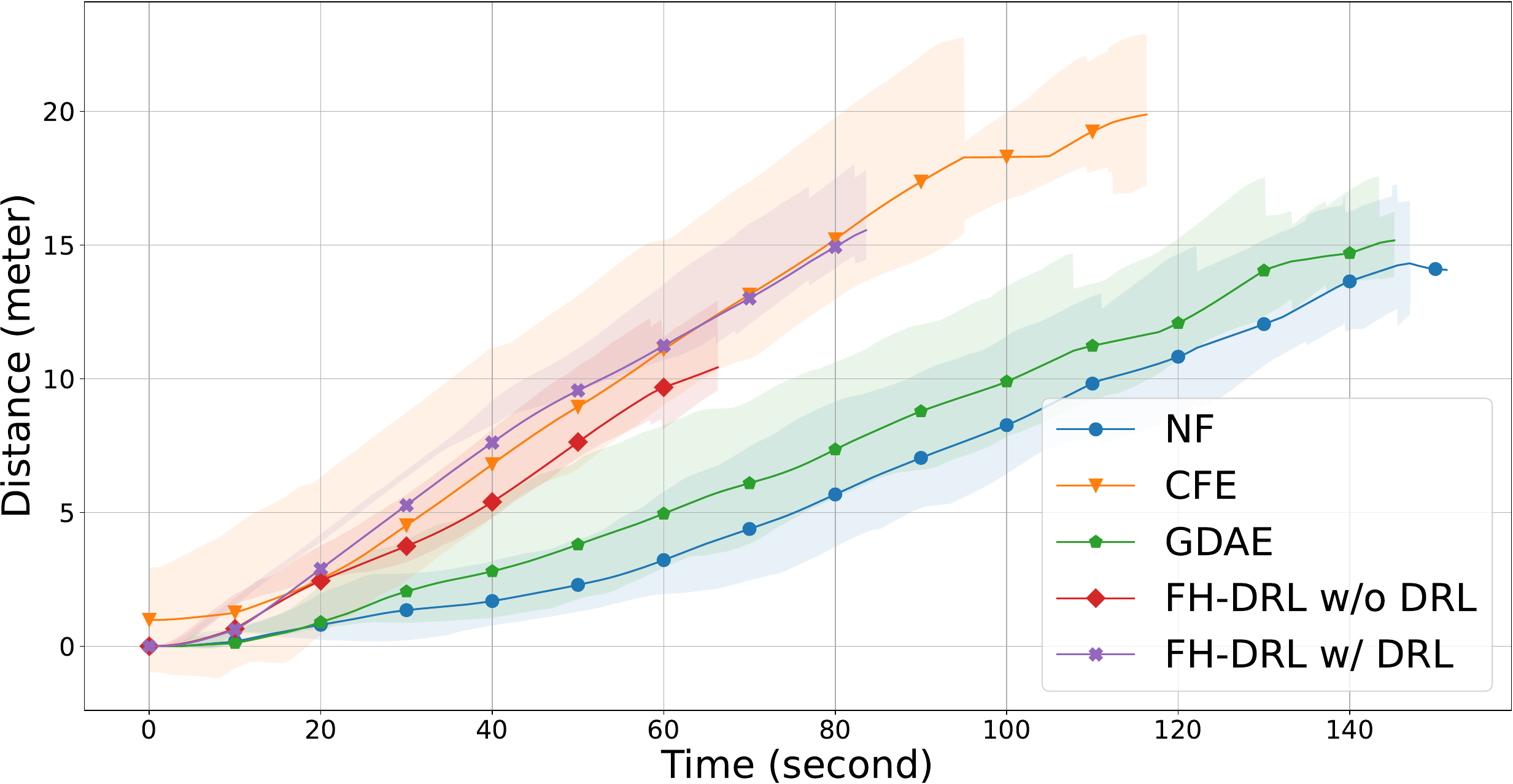}
		\caption{Scenario I (Distance by Time)}
		\label{fig_results_distance_by_time_world_01}
	\end{subfigure}
	\hfill
	\begin{subfigure}[b]{0.3\linewidth}
		\centering
		\includegraphics[width=1.0\linewidth]{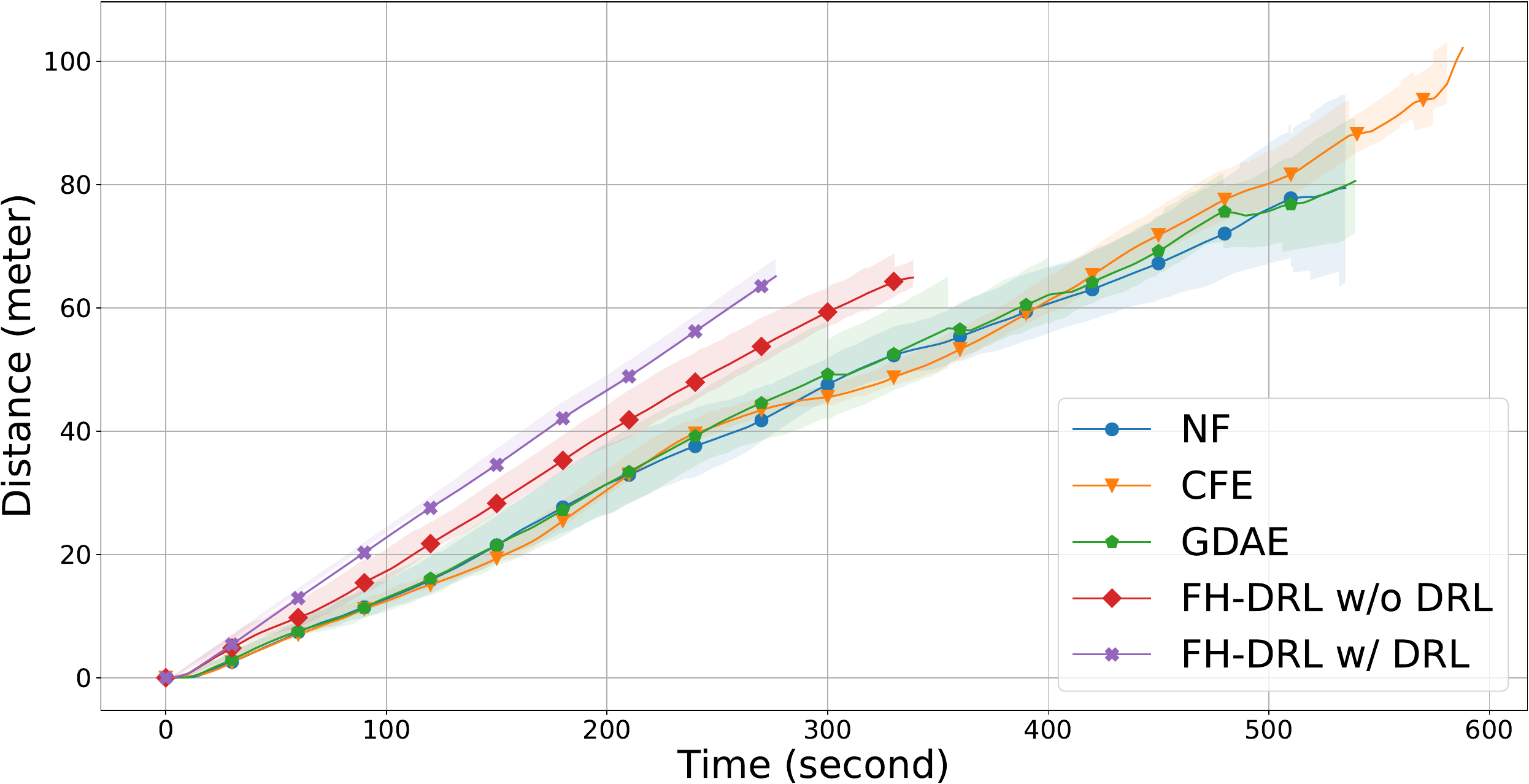}
		\caption{Scenario II (Distance by Time)}
		\label{fig_results_distance_by_time_world_02}
	\end{subfigure}
	\hfill
	\begin{subfigure}[b]{0.3\linewidth}
		\centering
		\includegraphics[width=1.0\linewidth]{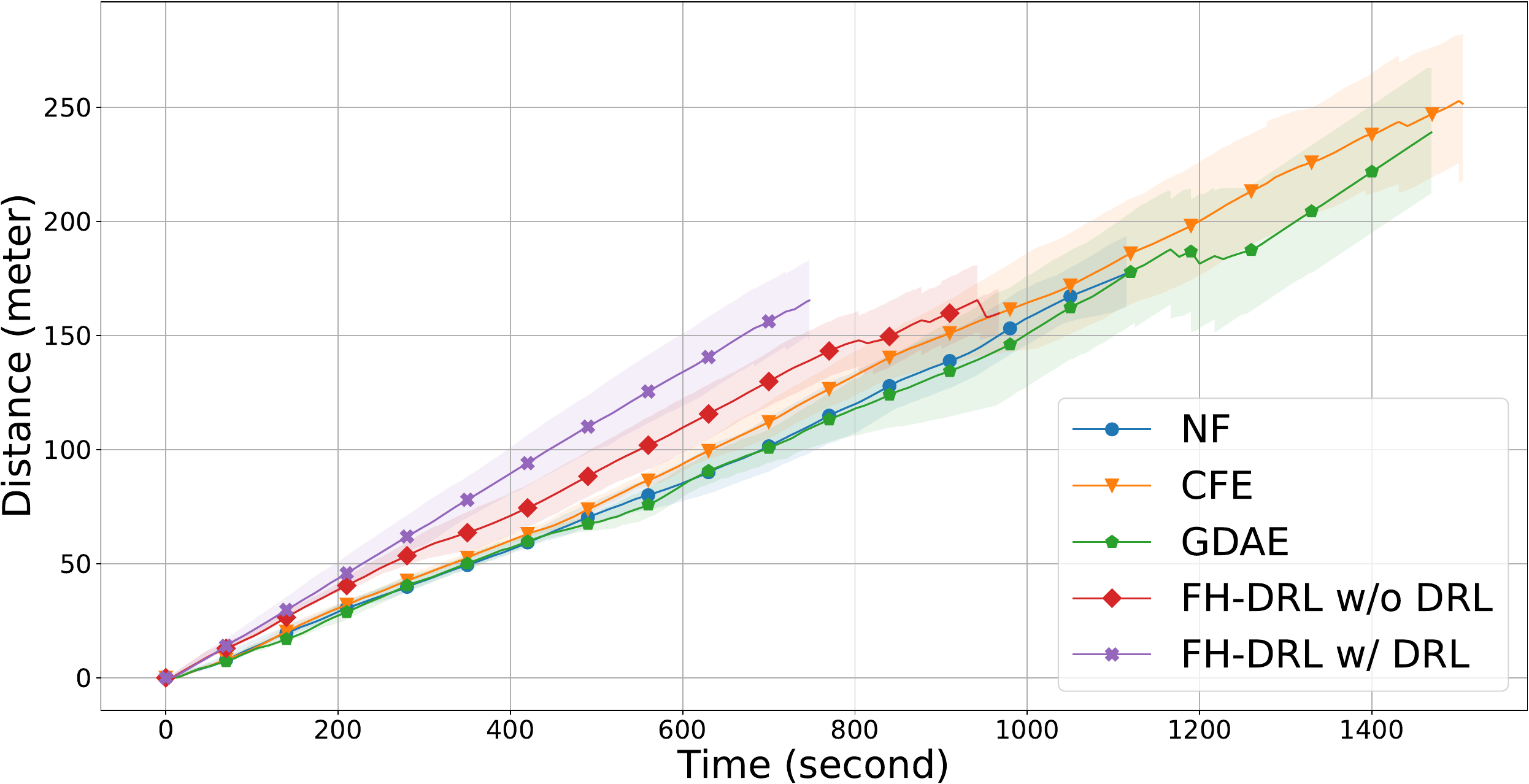}
		\caption{Scenario III (Distance by Time)}
		\label{fig_results_distance_by_time_world_03}
	\end{subfigure}

	\vfill
	\begin{subfigure}[b]{0.3\linewidth}
		\centering
		\includegraphics[width=1.0\linewidth]{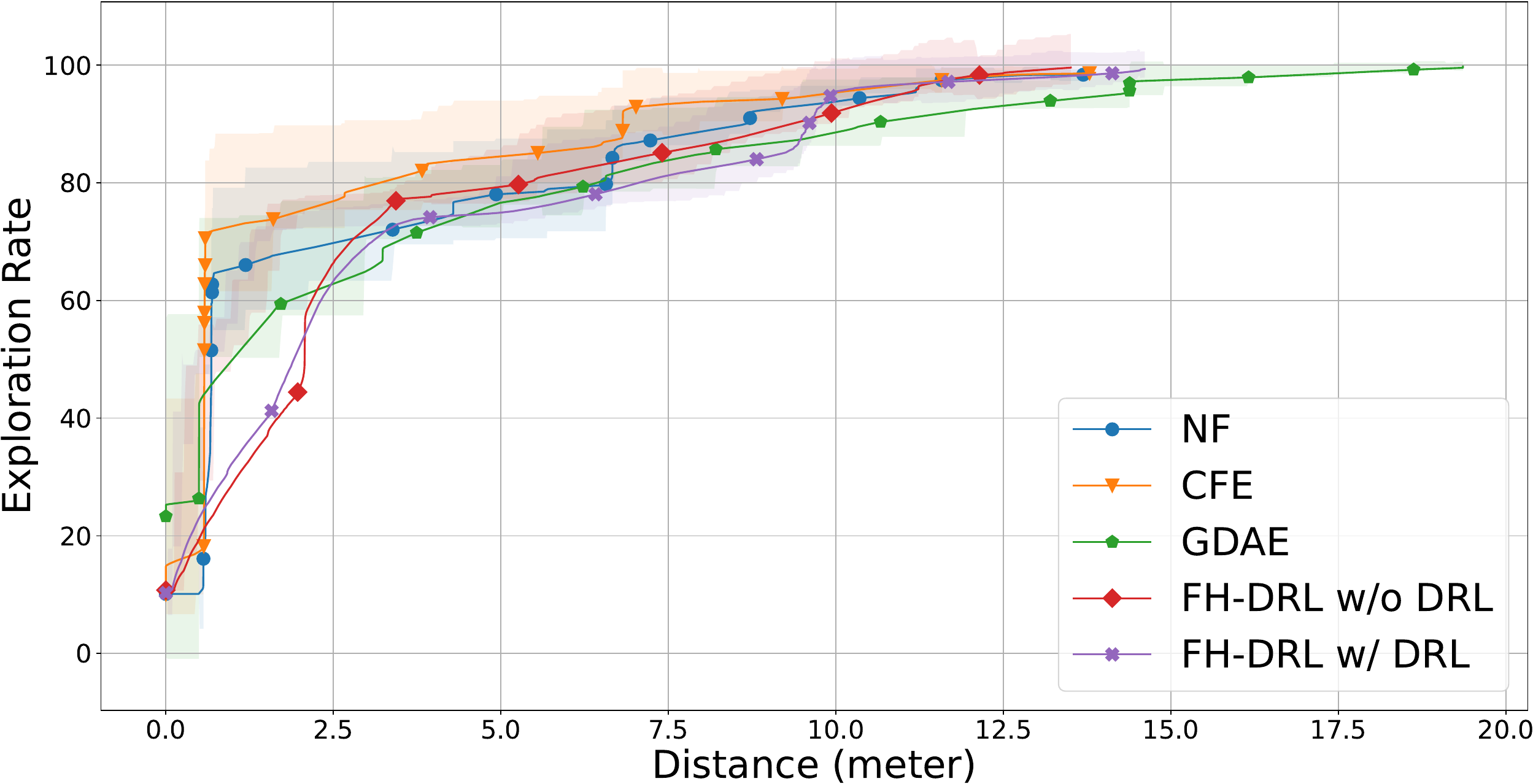}
		\caption{Scenario I (Rate by Distance)}
		\label{fig_results_rate_by_distance_world_01}
	\end{subfigure}
	\hfill
	\begin{subfigure}[b]{0.3\linewidth}
		\centering
		\includegraphics[width=1.0\linewidth]{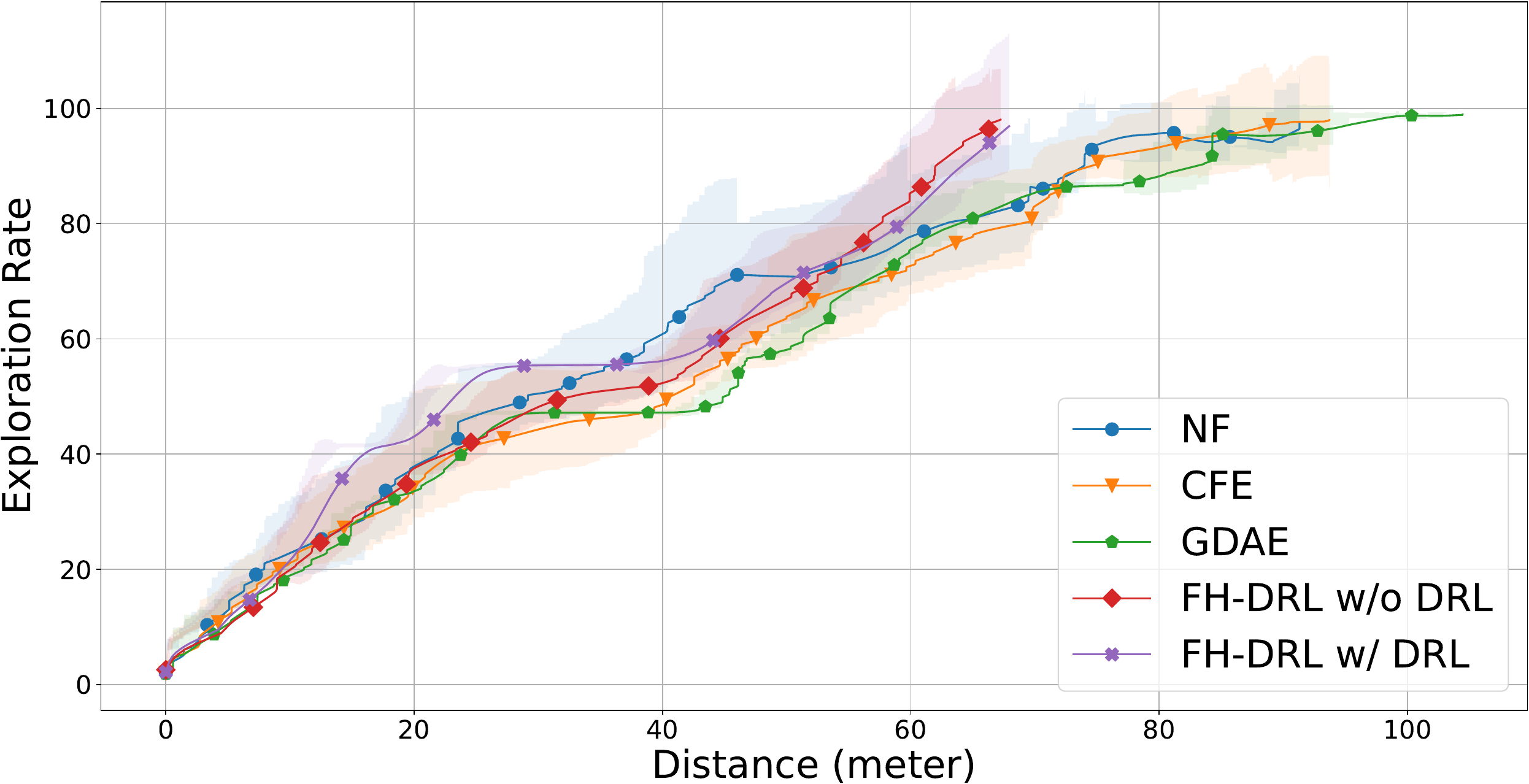}
		\caption{Scenario II (Rate by Distance)}
		\label{fig_results_rate_by_distance_world_02}
	\end{subfigure}
	\hfill
	\begin{subfigure}[b]{0.3\linewidth}
		\centering
		\includegraphics[width=1.0\linewidth]{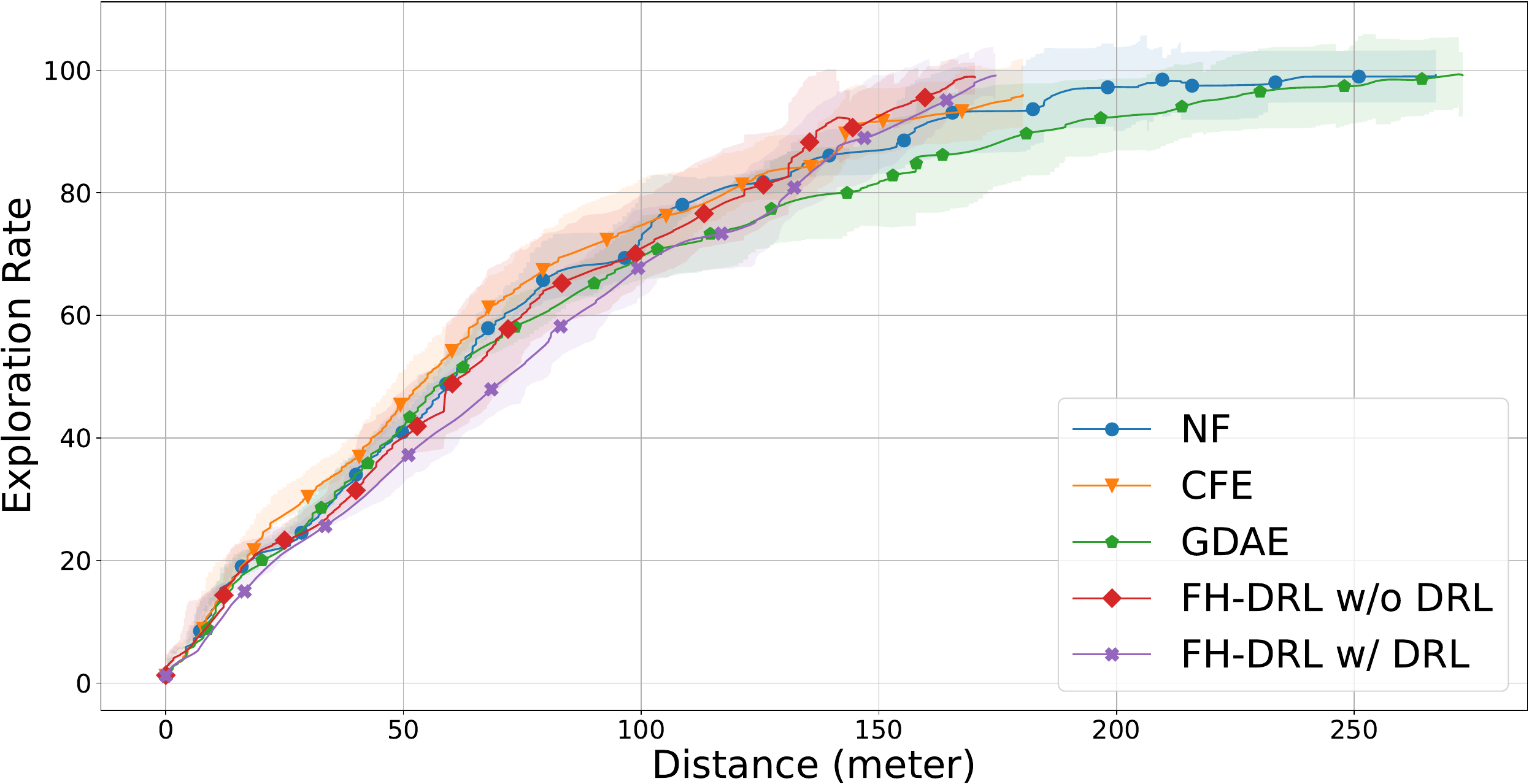}
		\caption{Scenario III (Rate by Distance)}
		\label{fig_results_rate_by_distance_world_03}
	\end{subfigure}

	\vfill
	\begin{subfigure}[b]{0.3\linewidth}
		\centering
		\includegraphics[width=1.0\linewidth]{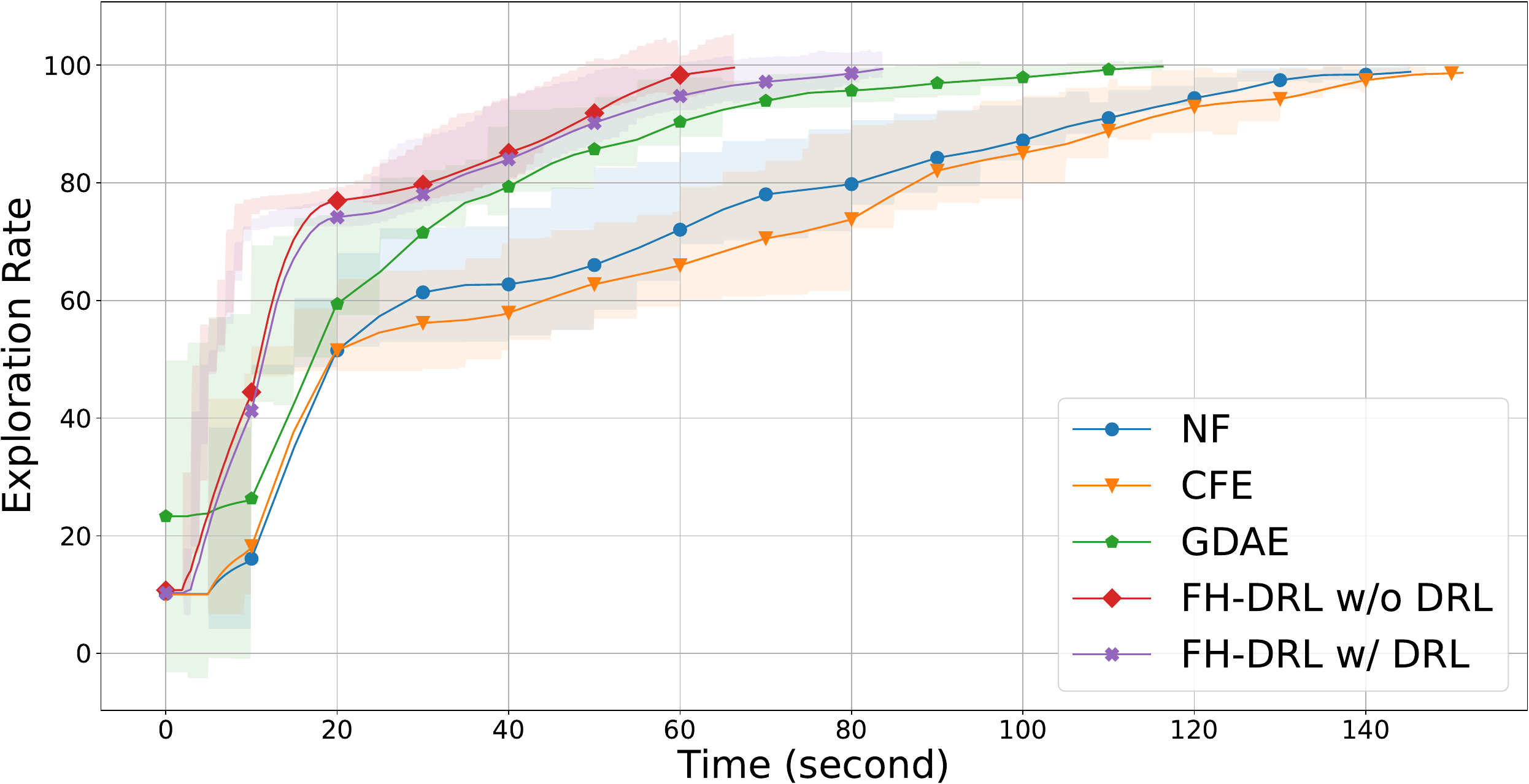}
		\caption{Scenario I (Rate by Time)}
		\label{fig_results_rate_by_time_world_01}
	\end{subfigure}
	\hfill
	\begin{subfigure}[b]{0.3\linewidth}
		\centering
		\includegraphics[width=1.0\linewidth]{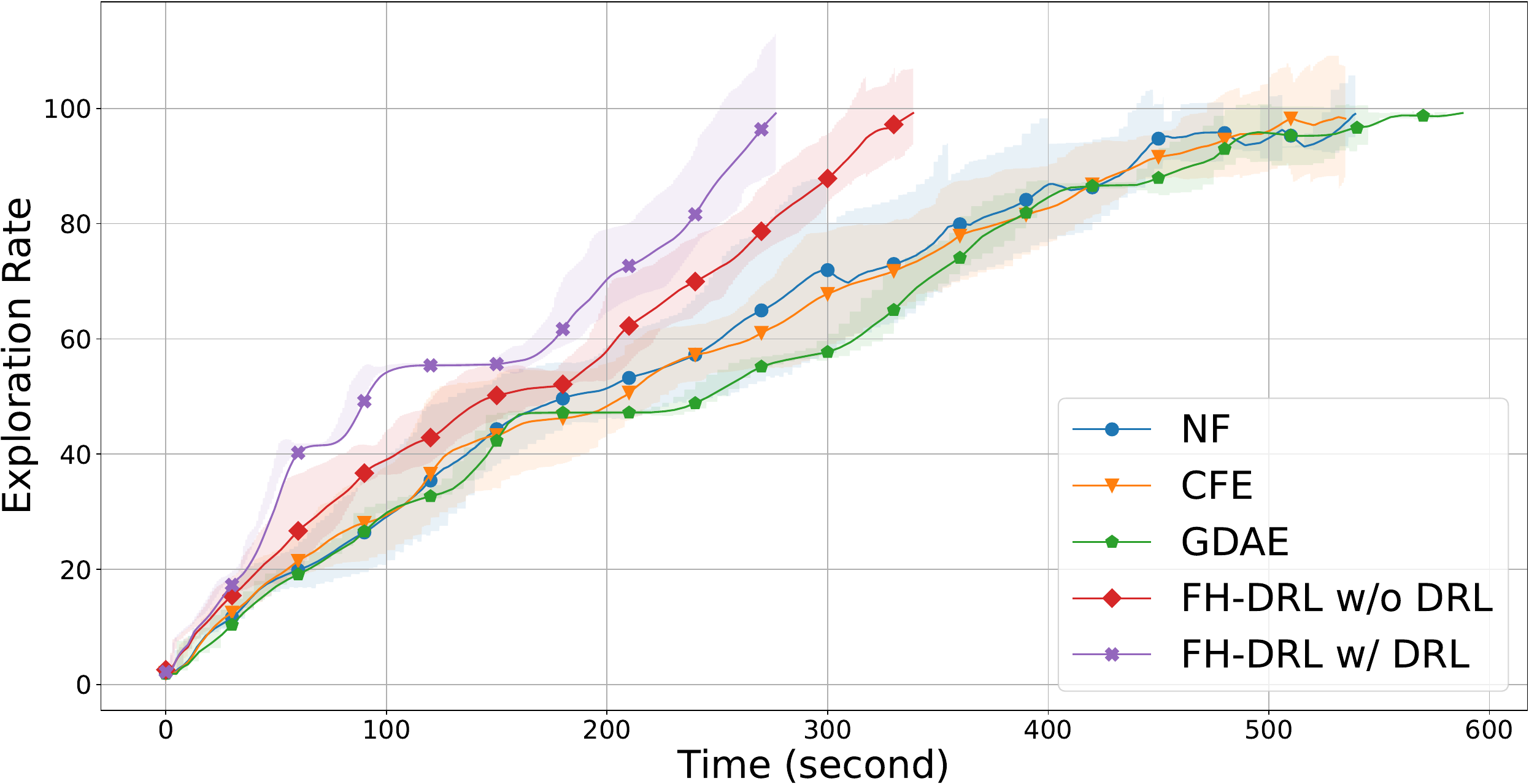}
		\caption{Scenario II (Rate by Time)}
		\label{fig_results_rate_by_time_world_02}
	\end{subfigure}
	\hfill
	\begin{subfigure}[b]{0.3\linewidth}
		\centering
		\includegraphics[width=1.0\linewidth]{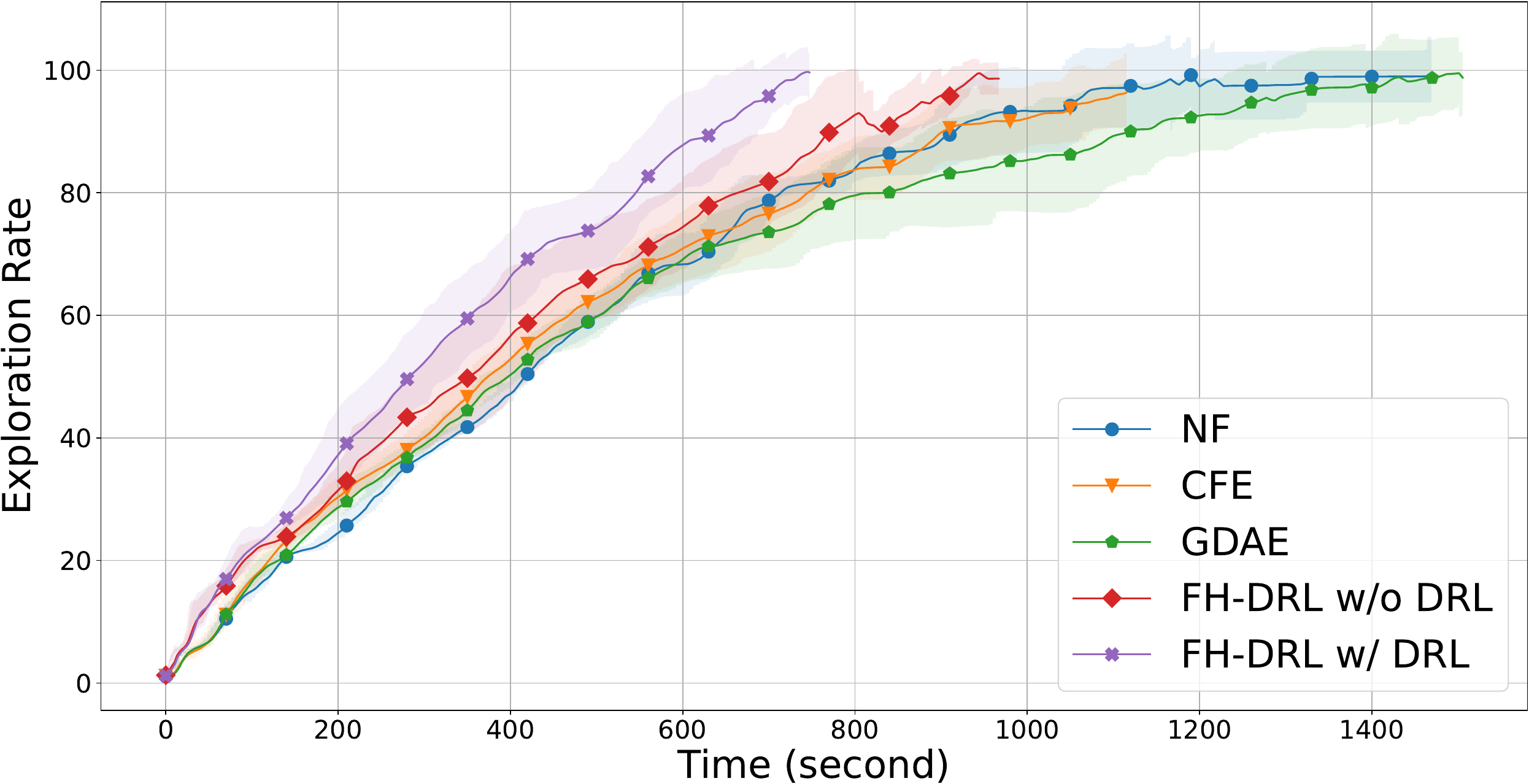}
		\caption{Scenario III (Rate by Time)}
		\label{fig_results_rate_by_time_world_03}
	\end{subfigure}

	\vfill
	\begin{subfigure}[b]{0.27\linewidth}
		\centering
		\includegraphics[width=1.0\linewidth]{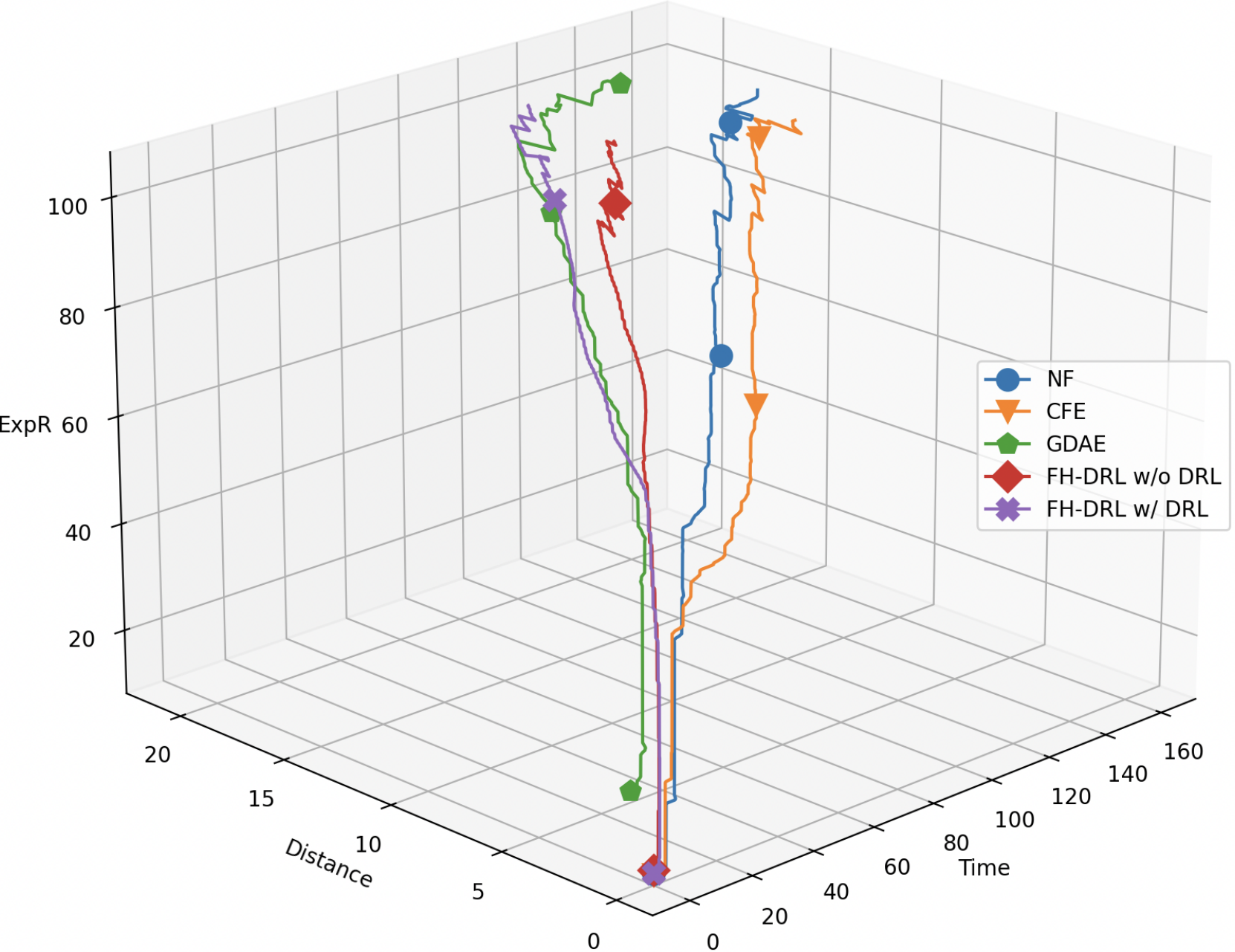}
		\caption{Scenario I (3D)}
		\label{fig_3d_results_world1}
	\end{subfigure}
	\hfill
	\begin{subfigure}[b]{0.27\linewidth}
		\centering
		\includegraphics[width=1.0\linewidth]{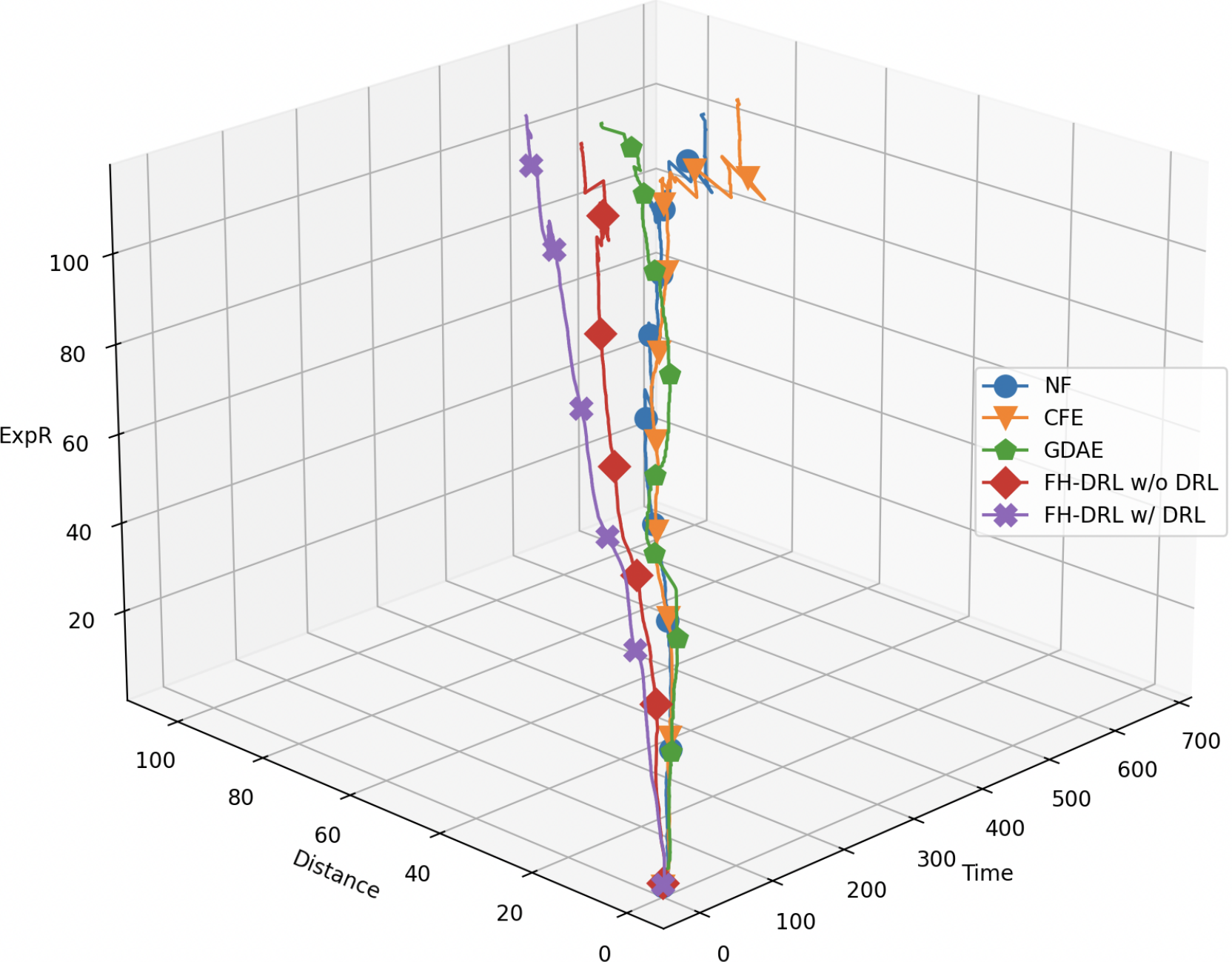}
		\caption{Scenario II (3D)}
		\label{fig_3d_results_world2}
	\end{subfigure}
	\hfill
	\begin{subfigure}[b]{0.27\linewidth}
		\centering
		\includegraphics[width=1.0\linewidth]{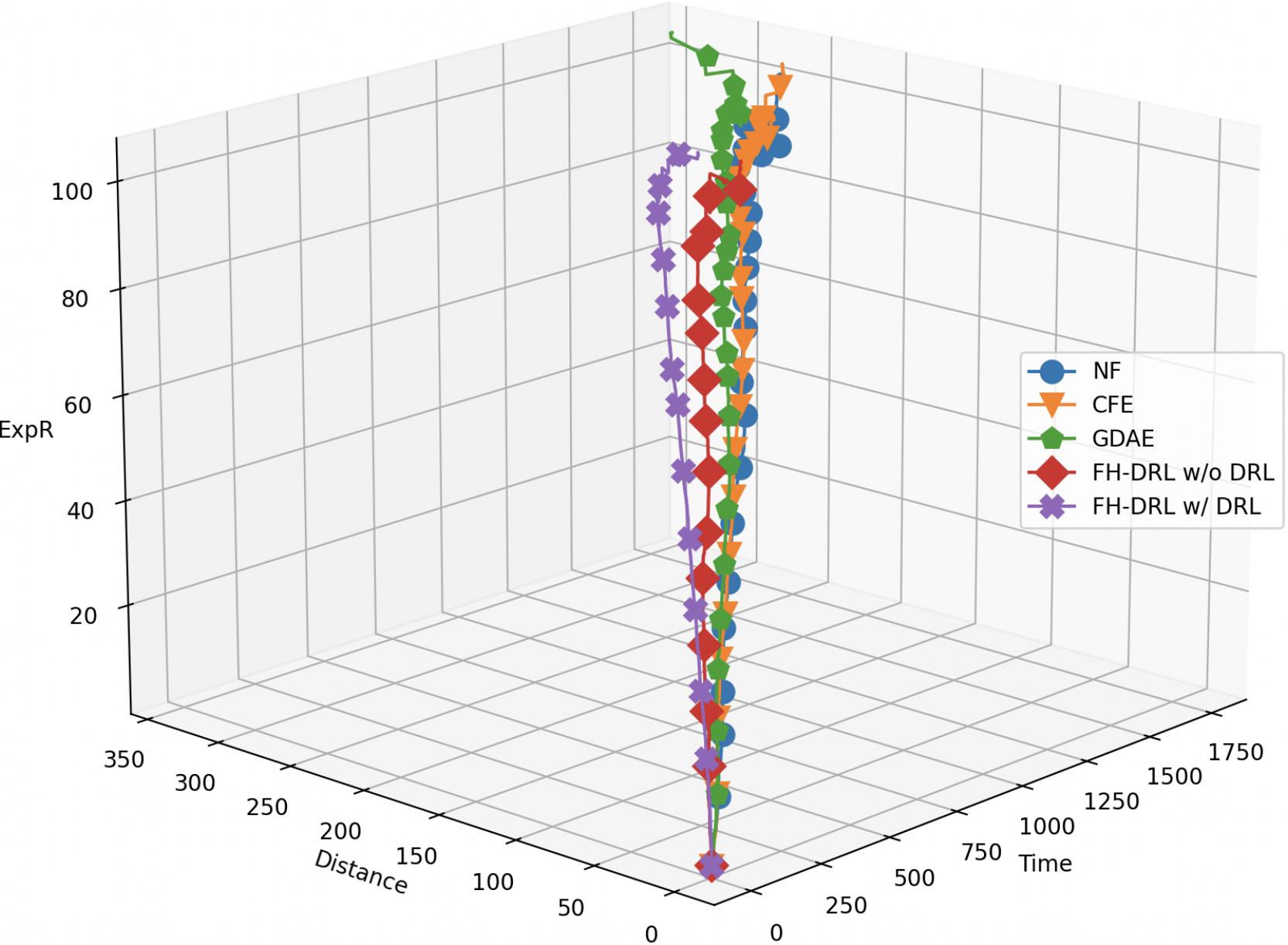}
		\caption{Scenario III (3D)}
		\label{fig_3d_results_world3}
	\end{subfigure}

	\caption{Experimental Simulation Results of Robot Exploration in Unknown Spaces}
	\label{fig_experimental_simulation_results}
\end{figure*}

\begin{figure}
	\centering
	\includegraphics[width=1\linewidth]{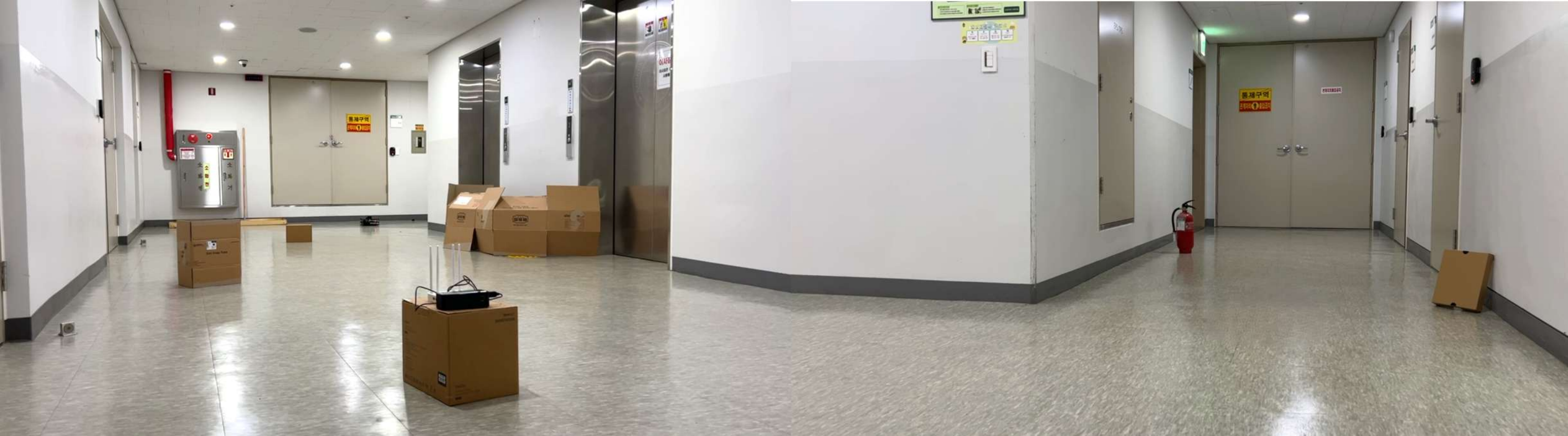}
	\caption{Real World Experiment Environment}
	\label{fig_real_world_experiment_environment}
\end{figure}

\subsubsection{Distance by Time}
Figures~\ref{fig_results_distance_by_time_world_01}, \ref{fig_results_distance_by_time_world_02}, and \ref{fig_results_distance_by_time_world_03} depict the distance traveled as a function of time for GDAE, NF, CFE, FH-DRL with DRL, and FH-DRL without DRL across unknown environments (Scenarios~I, II, and III). The shaded regions represent variance, reflecting the reliability of each method. In Scenario~I, FH-DRL with DRL attains the lowest distances and completion times, closely followed by FH-DRL without DRL. CFE exhibits moderate performance, while NF and GDAE demonstrate less efficiency. As complexity increases (Scenarios~II and III), FH-DRL with DRL continues to outperform its counterparts by completing tasks more rapidly and efficiently. Overall, FH-DRL consistently provides superior performance—particularly in complex environments—emphasising its effectiveness in exploration tasks.

\subsubsection{Exploration Rate by Distance}
Figures~\ref{fig_results_rate_by_distance_world_01}, \ref{fig_results_rate_by_distance_world_02}, and \ref{fig_results_rate_by_distance_world_03} illustrate the relationship between exploration rate and distance. In Scenario~I, NF achieves a high exploration rate over relatively short distances. In the more challenging Scenarios~II and III, FH-DRL (both with and without DRL) demonstrates a noticeable performance boost and surpasses other methods. Specifically, in Scenario~III, FH-DRL with DRL proves the most efficient, achieving high exploration rates with minimal traveled distance. By contrast, CFE, GDAE, and NF display lower performance, while FH-DRL excels in more intricate environments.

\subsubsection{Exploration Rate by Time}
Figures illustrating exploration rate over time for the five approaches in Scenarios~I, II, and III reveal that FH-DRL with DRL swiftly reaches approximately 80\% coverage within the first 20 seconds in Scenario~I, followed by FH-DRL without DRL. In Scenarios~II and III, FH-DRL with DRL maintains its advantage by achieving higher exploration rates in shorter time spans, again outpacing FH-DRL without DRL. CFE demonstrates moderate progress, while NF and GDAE lag behind. Taken together, these results indicate that integrating DRL substantially enhances the exploration efficiency of FH-DRL.

\subsubsection{Exploration Rate by Distance and Time}
Figures~\ref{fig_3d_results_world1}--\ref{fig_3d_results_world3} present three-dimensional plots relating exploration rate to both distance and time. FH-DRL (with and without DRL) not only covers greater distances in less time but also maintains high exploration rates in all scenarios. In contrast, GDAE, NF, and CFE face challenges over longer distances and durations, whereas FH-DRL consistently achieves more rapid and comprehensive exploration.

\subsubsection{Traversed Paths}
\begin{figure*}[!htbp]
	\centering
	\begin{subfigure}[b]{0.25\linewidth}
		\centering
		\includegraphics[width=0.80\linewidth]{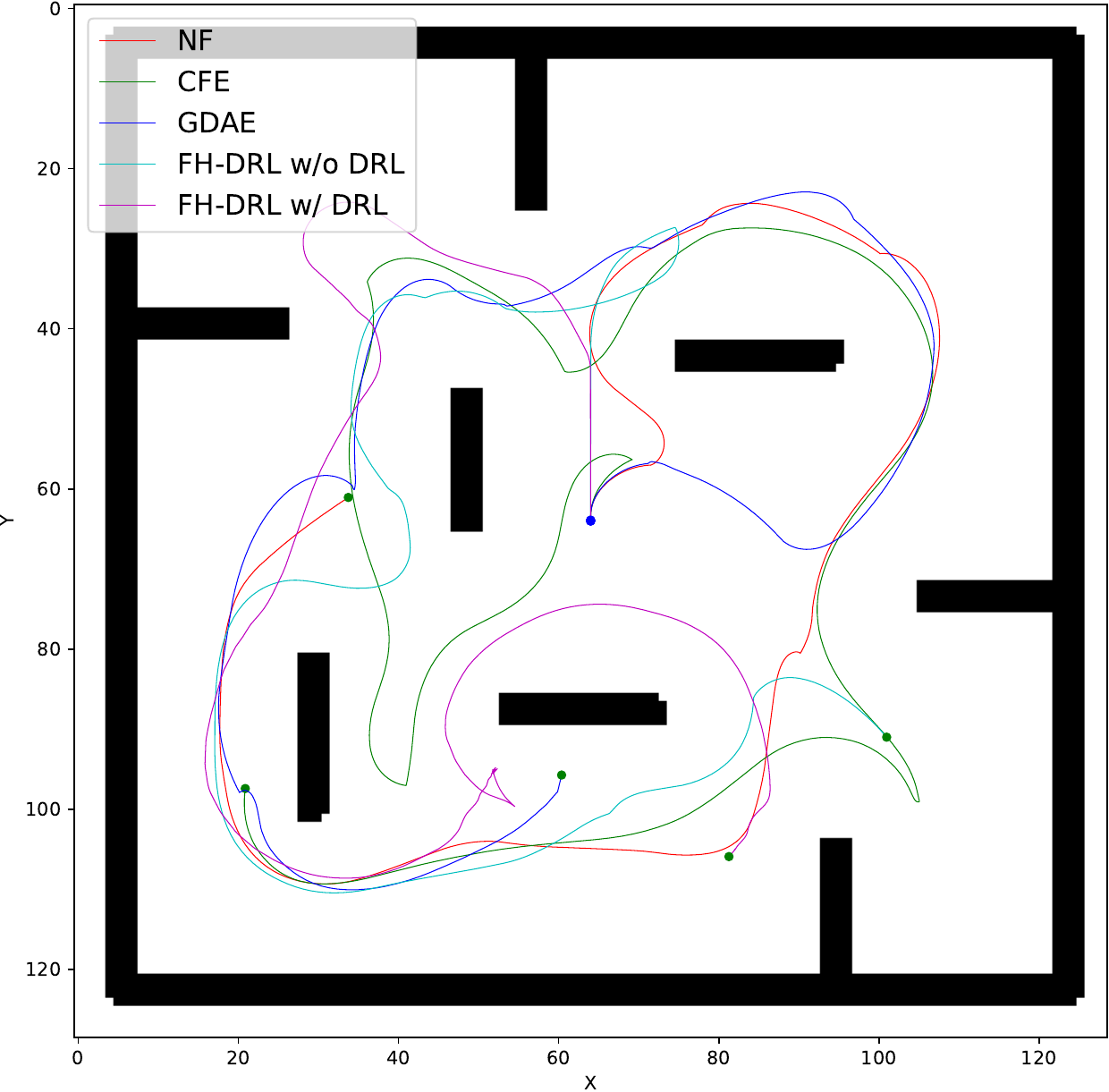}
		\caption{Scenario I}
		\label{fig_results_exploration_path_world_01}
	\end{subfigure}
	\hfill
	\begin{subfigure}[b]{0.335\linewidth}
		\centering
		\includegraphics[width=1.0\linewidth]{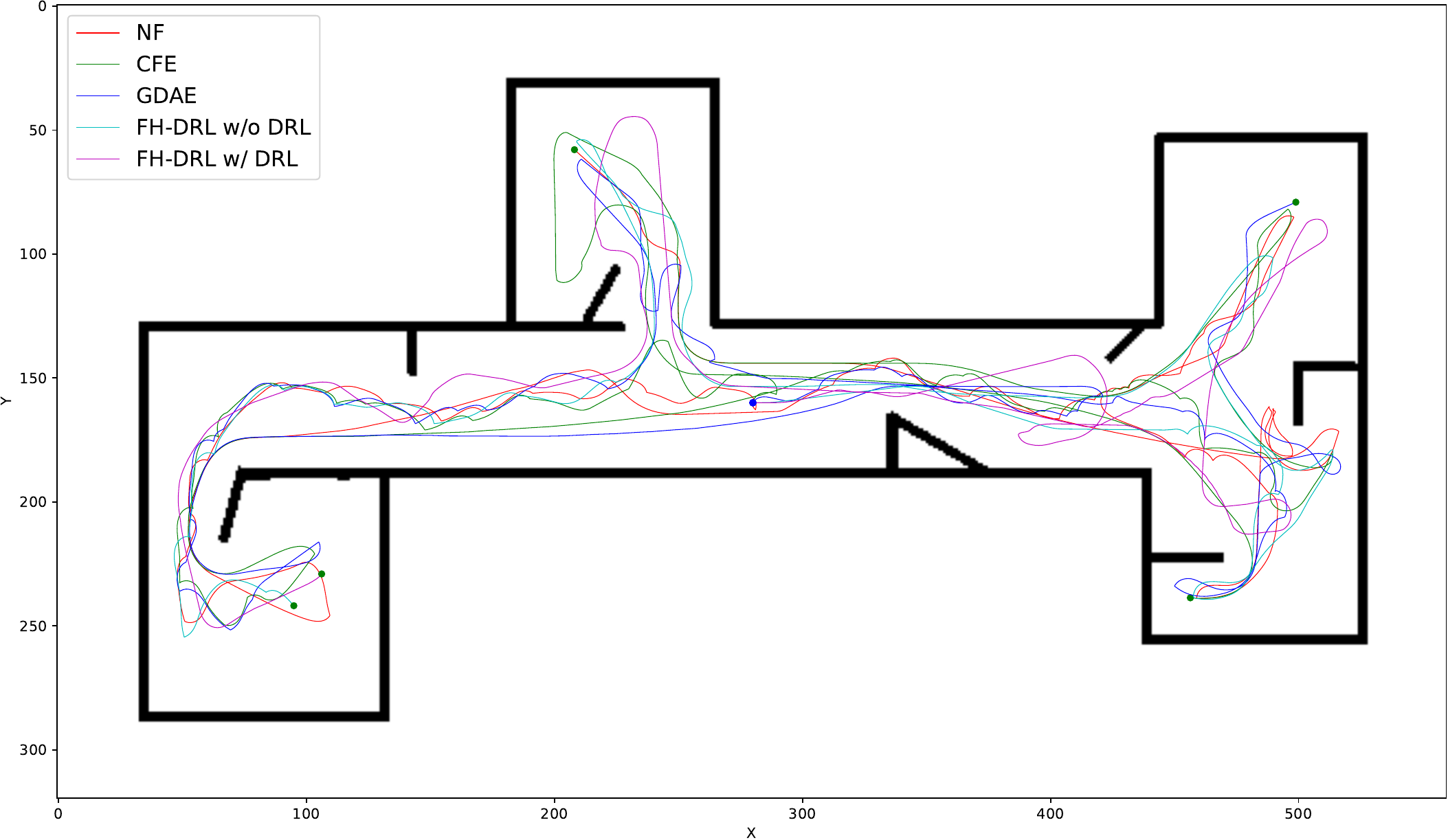}
		\caption{Scenario II}
		\label{fig_results_exploration_path_world_02}
	\end{subfigure}
	\hfill
	\begin{subfigure}[b]{0.395\linewidth}
		\centering
		\includegraphics[width=1.0\linewidth]{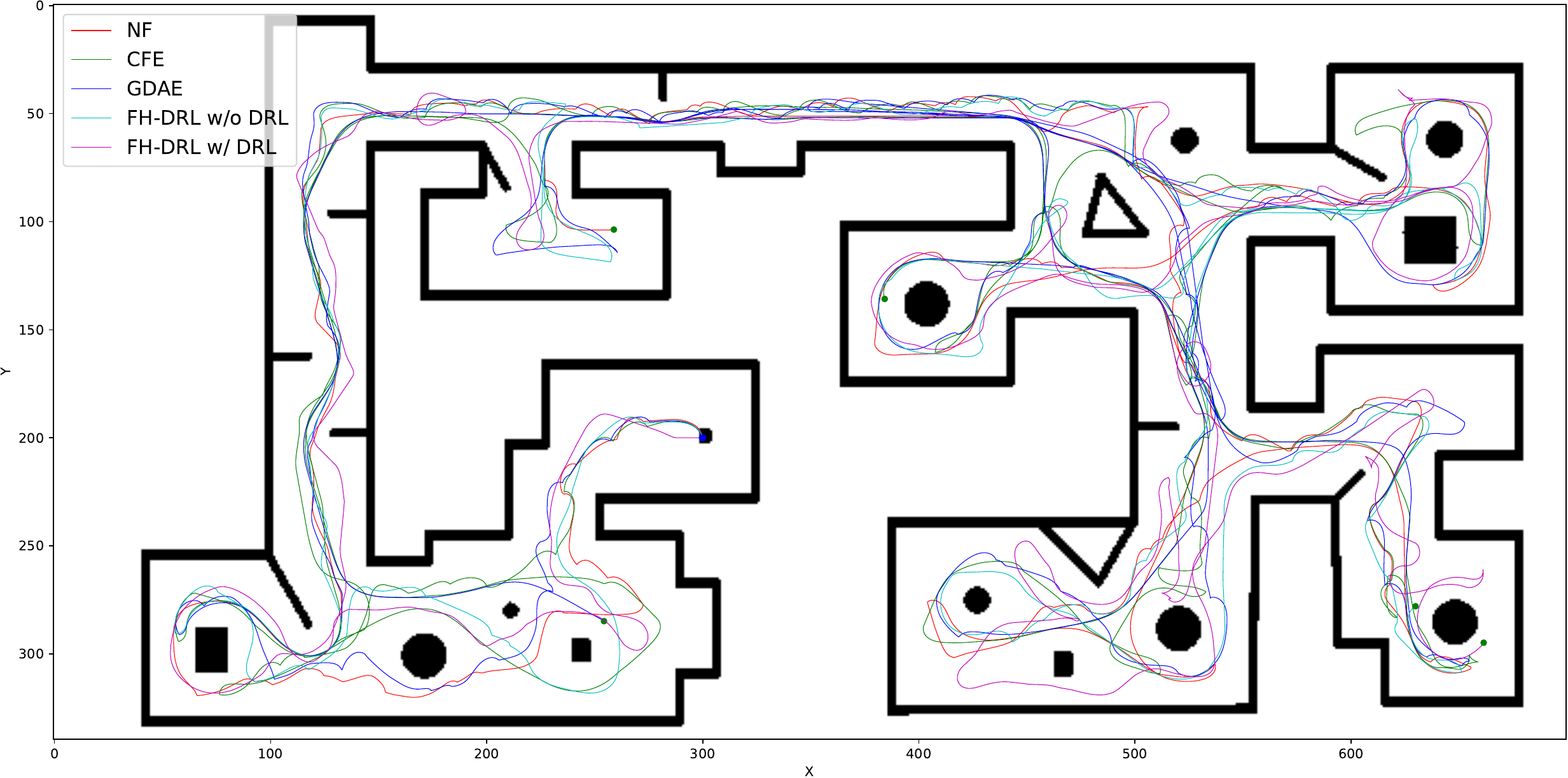}
		\caption{Scenario III}
		\label{fig_results_exploration_path_world_03}
	\end{subfigure}
	\caption{Exploration paths}
	\label{fig_results_exploration_path_world}
\end{figure*}

Figure~\ref{fig_results_exploration_path_world} illustrates the robot trajectories for each method across three map complexities. FH-DRL with DRL attains the most efficient paths, characterized by minimal redundancy, while FH-DRL without DRL shows moderate retracing. NF and CFE present multiple overlapping routes, and although GDAE performs better than these two, it still falls short of FH-DRL. Notably, the DRL-based FH-DRL exhibits the highest path efficiency, especially under Scenario~III.


\subsubsection{Discussion}
FH-DRL without DRL produces the shortest average travel distances, whereas FH-DRL with DRL attains the lowest average completion times, indicating more rapid exploration. Both variants maintain consistently high exploration rates with minimal variance. Leveraging both heuristic frontier selection and DRL-based navigation proves highly effective in unknown environments, outperforming competing approaches that depend on \texttt{nav2}-based planning and discrete movement. By integrating continuous navigation through DRL with a heuristic frontier-selection mechanism, FH-DRL achieves comprehensive map coverage in reduced time, thereby demonstrating particular promise for autonomous exploration tasks.

\begin{figure}
	\centering
	\begin{subfigure}[b]{0.48\linewidth}
		\centering
		\includegraphics[width=1.0\linewidth]{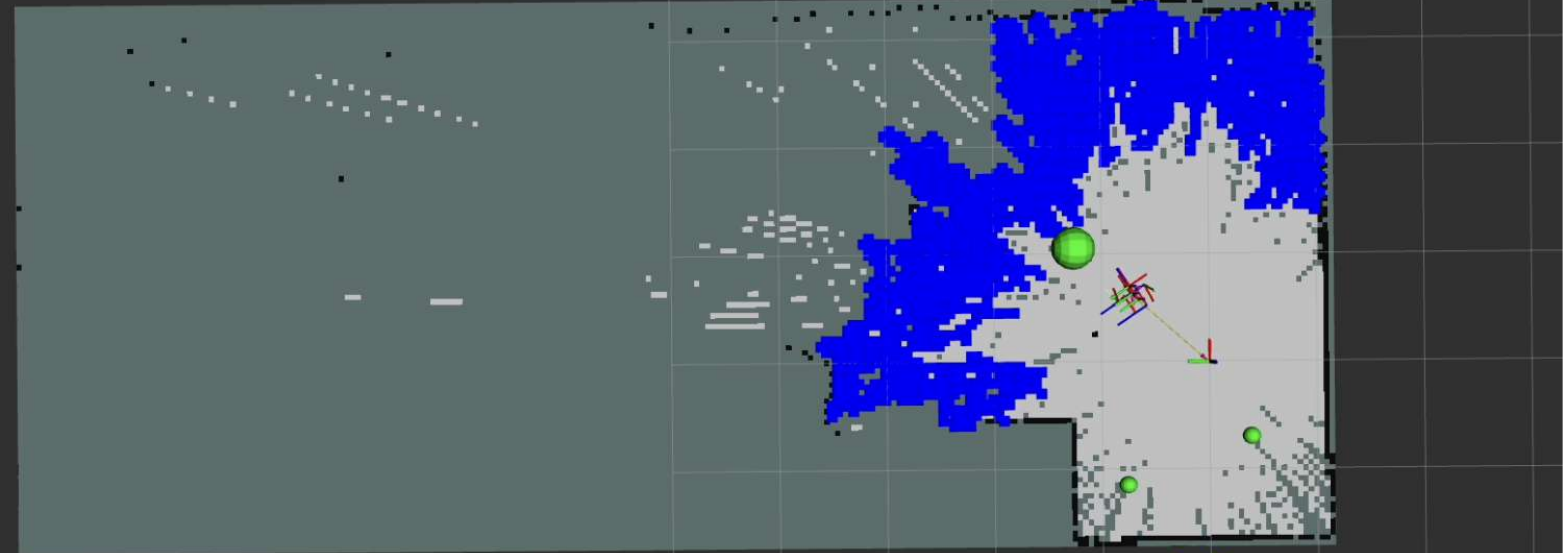}
		\caption{Real World Scenario Step I}
		\label{fig_real_world_experiment_0}
	\end{subfigure}
	\hfill
	\begin{subfigure}[b]{0.48\linewidth}
		\centering
		\includegraphics[width=1.0\linewidth]{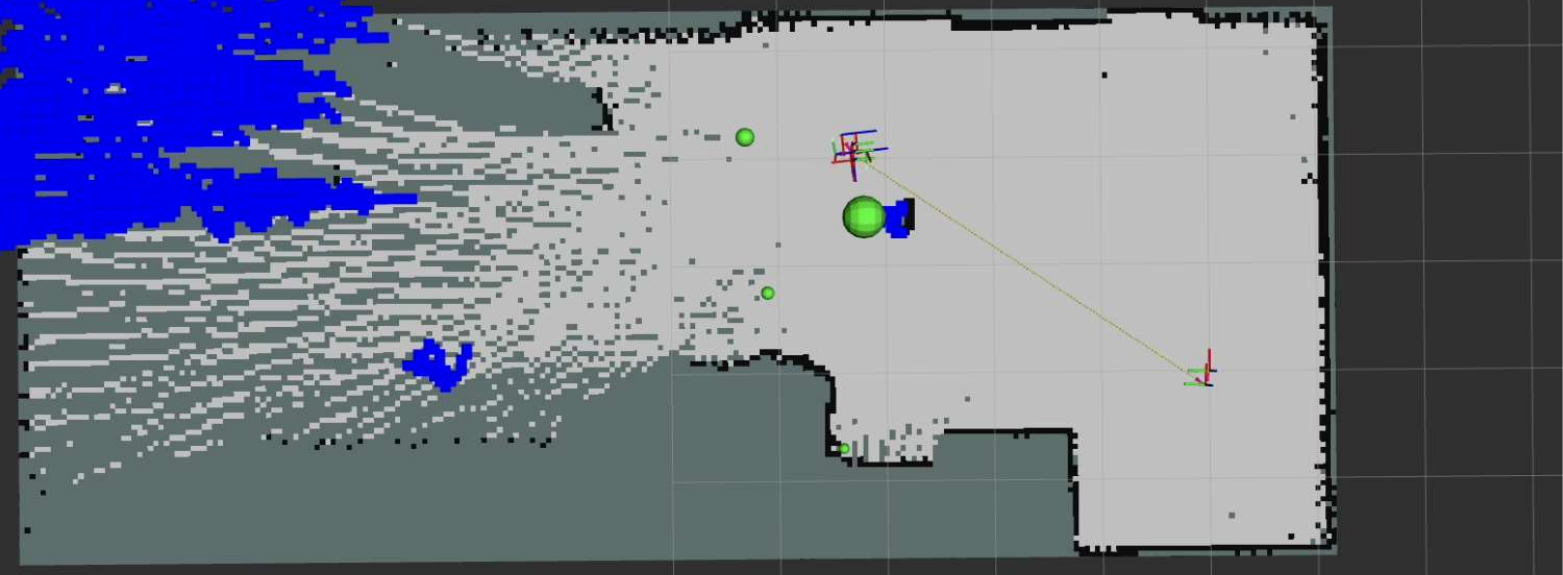}
		\caption{Real World Scenario Step II}
		\label{fig_real_world_experiment_1}
	\end{subfigure}

	\vfill
	\begin{subfigure}[b]{0.48\linewidth}
		\centering
		\includegraphics[width=1.0\linewidth]{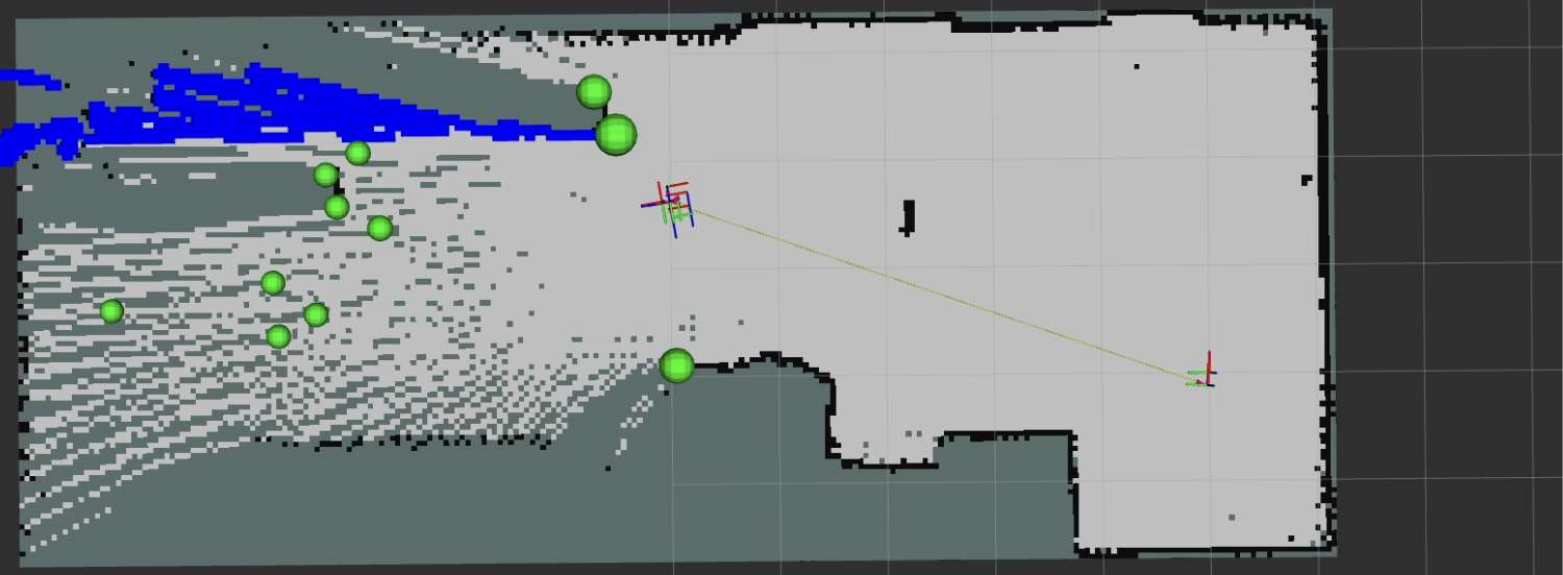}
		\caption{Real World Scenario Step III}
		\label{fig_real_world_experiment_2}
	\end{subfigure}
	\hfill
	\begin{subfigure}[b]{0.48\linewidth}
		\centering
		\includegraphics[width=1.0\linewidth]{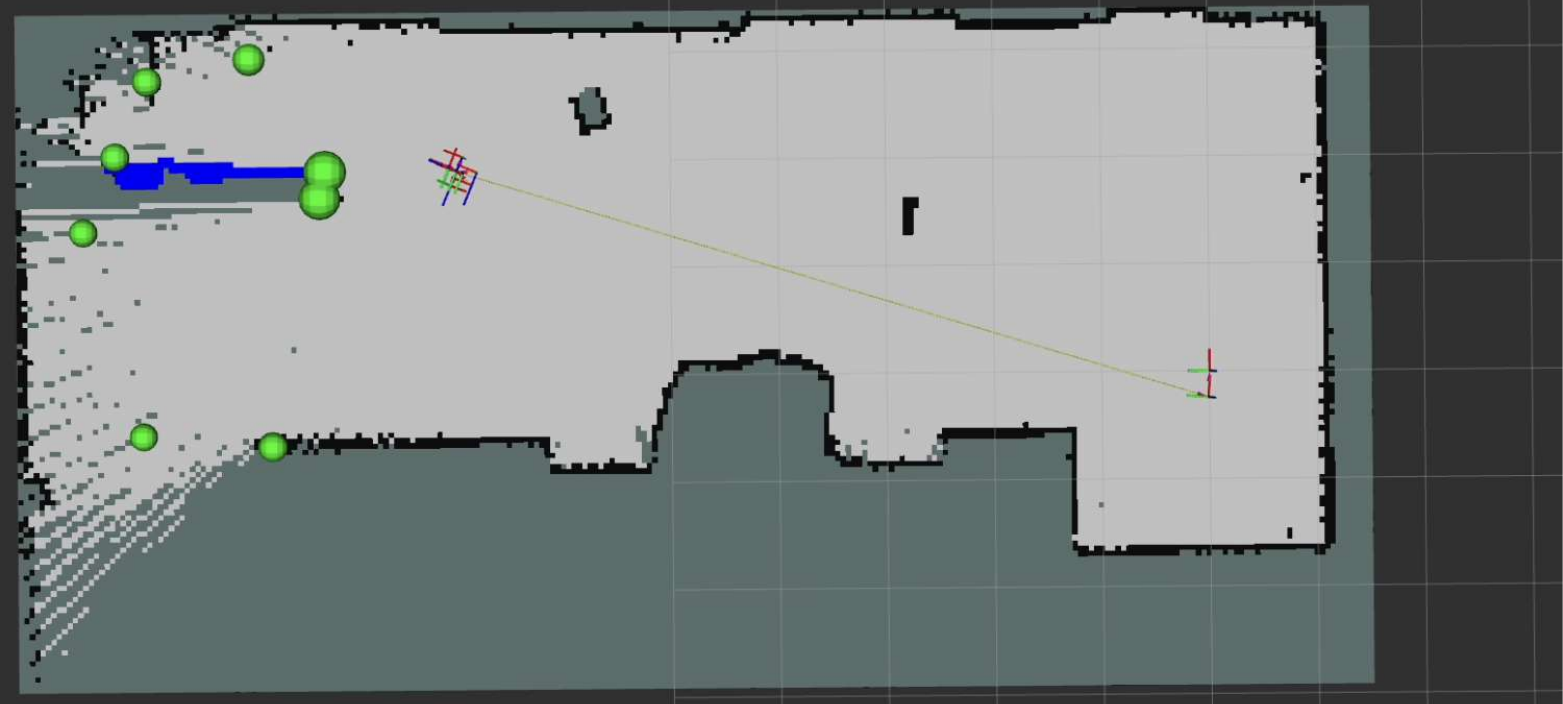}
		\caption{Real World Scenario Step IV}
		\label{fig_real_world_experiment_3}
	\end{subfigure}

	\vfill
	\begin{subfigure}[b]{0.48\linewidth}
		\centering
		\includegraphics[width=1.0\linewidth]{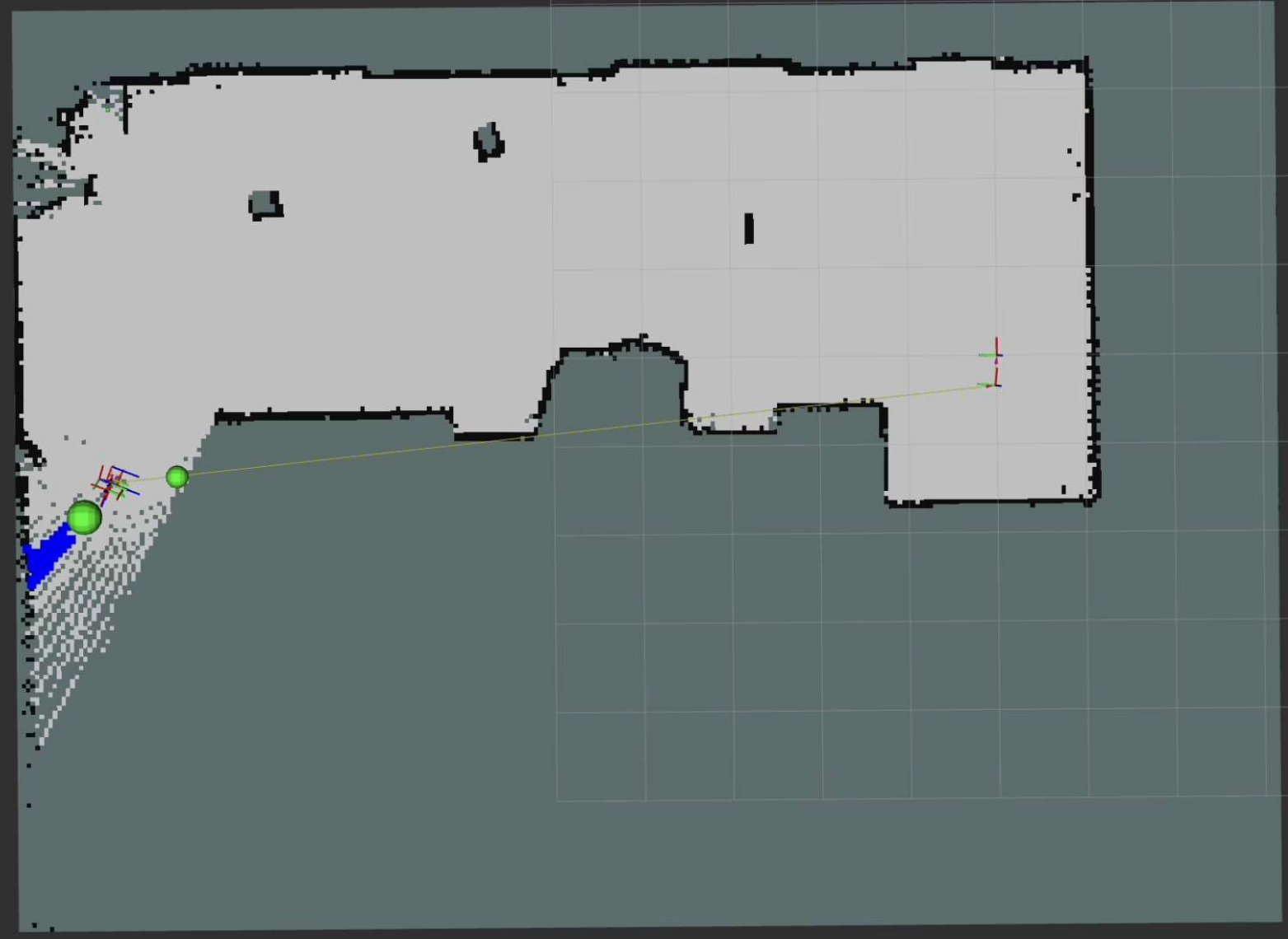}
		\caption{Real World Scenario Step V}
		\label{fig_real_world_experiment_4}
	\end{subfigure}
	\hfill
	\begin{subfigure}[b]{0.48\linewidth}
		\centering
		\includegraphics[width=1.0\linewidth]{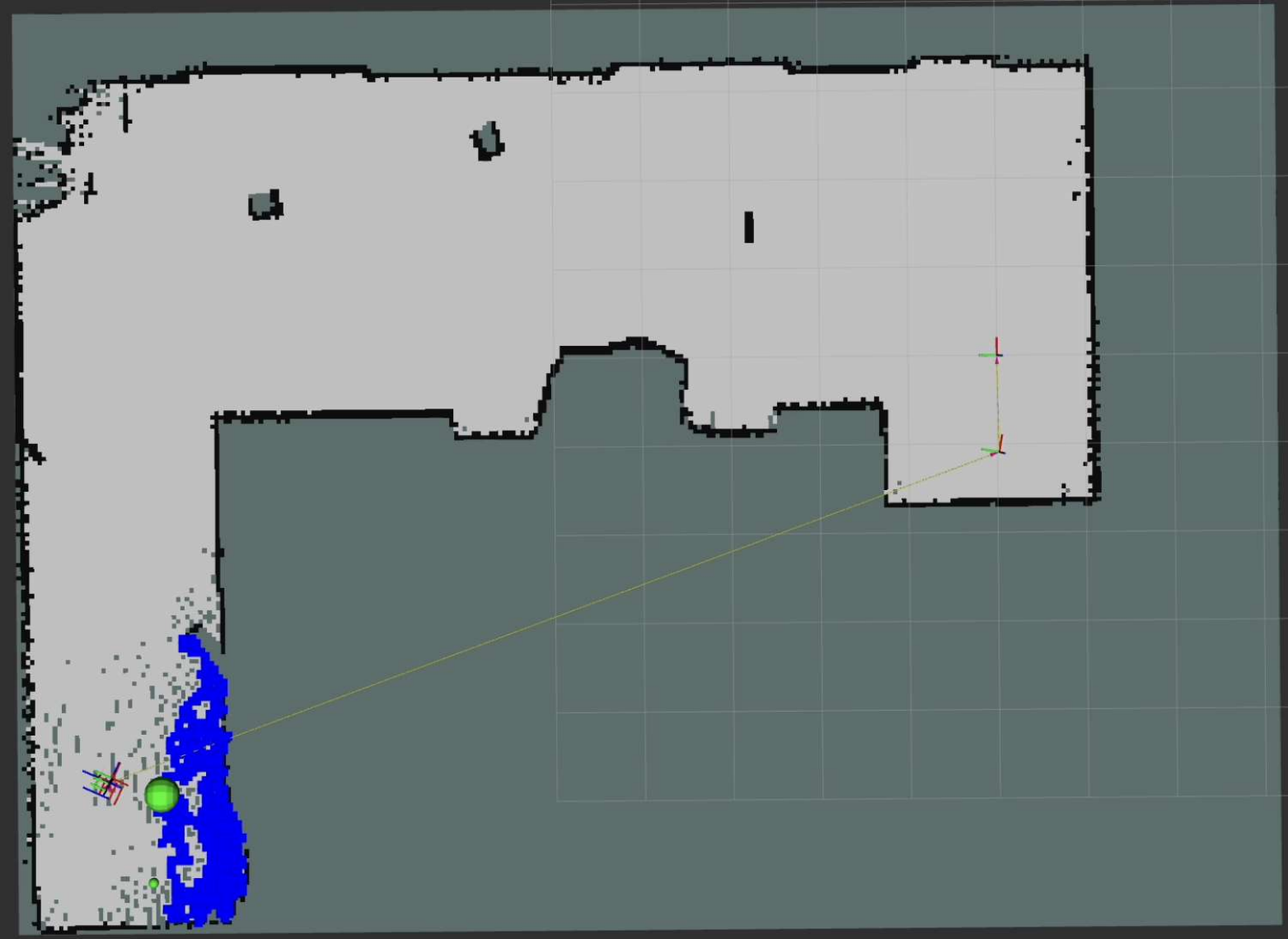}
		\caption{Real World Scenario Step VI}
		\label{fig_real_world_experiment_5}
	\end{subfigure}

	\vfill
	\begin{subfigure}[b]{0.8\linewidth}
		\centering
		\includegraphics[width=1.0\linewidth]{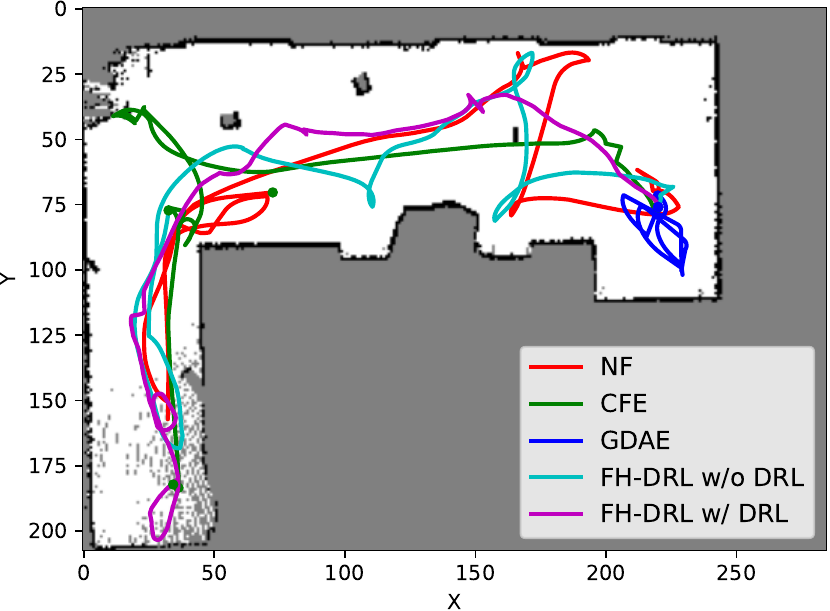}
		\caption{Trajactories}
		\label{fig_real_world_experiment_trajactories}
	\end{subfigure}

	\caption{Real World Experiment Phases in Unknown Spaces}
	\label{fig_real_experiment_results}
\end{figure}

\subsection{Real-World Experiments}

A real-world evaluation was performed using a Turtlebot3 \texttt{waffle\_pi} in a structured indoor hallway, as shown in Figure~\ref{fig_real_world_experiment_environment}. This environment includes both straight and cornered sections, along with randomly placed box obstacles, thereby testing the robot’s navigation and decision-making capabilities. The narrow corridor junction further challenges path-planning algorithms, while the confined layout and features provide a rigorous testbed for the proposed FH-DRL algorithm under realistic conditions. The experimental progression is illustrated in Figure~\ref{fig_real_experiment_results}, highlighting the frontier-exploration process in an actual setting.

In Figure~\ref{fig_real_world_experiment_0}, the algorithm initially selects the largest frontier based on both proximity and $\mathcal{O}(x_f, y_f)$. Figure~\ref{fig_real_world_experiment_1} demonstrates the selection of a \textbf{closed frontier} which might otherwise have been overlooked if frontier size alone were considered. Figure~\ref{fig_real_world_experiment_2} underscores a preference for an \textbf{open wide frontier} over a nearer \textbf{door gap} frontier, thereby expanding exploration opportunities. In Figure~\ref{fig_real_world_experiment_3}, a frontier is chosen based on $\mathcal{D}(x_f, y_f)$ to minimise exploration time, even though an alternative frontier has a lower $\mathcal{O}(x_f, y_f)$. Figure~\ref{fig_real_world_experiment_4} highlights the robot’s decision to pursue a \textbf{closed frontier}, ensuring full coverage while avoiding the need to revisit the area in the future. Lastly, Figure~\ref{fig_real_world_experiment_5} indicates that the exploration has been successfully completed.

\begin{table}[h!]
	\centering
	\caption{Performance Comparison on Real World}
	\label{tab_performance_comparison}
	\begin{tabular}{lcc}
	\toprule
	\textbf{Method} & \textbf{Time (s)} & \textbf{Distance (m)} \\
	\midrule
	NF                & 298.4  & 32.2237 \\
	CFE               & 423.2  & 33.7278 \\
	GDAE              & 179.5  & 24.5279 \\
	FH-DRL w/o DRL  & \textcolor{blue}{172.1}  & \textcolor{blue}{23.8355} \\
	FH-DRL w/ DRL   & \textcolor{red}{132.3}  & \textcolor{red}{22.85} \\
	\bottomrule
	\end{tabular}
\end{table}

The corresponding experimental results, displayed in Figure~\ref{fig_real_world_experiment_trajactories} and Table~\ref{tab_performance_comparison}, illustrate that \textbf{FH-DRL with DRL} surpasses all other frontier-based exploration algorithms in both time and distance metrics.

In terms of time efficiency, \emph{FH-DRL with DRL} achieves the shortest exploration duration at \textbf{132.3\,s}, significantly outperforming its nearest competitor, FH-DRL without DRL, and far exceeding traditional methods such as CFE and NF. In parallel, \emph{FH-DRL with DRL} also records the shortest travel distance (\textbf{22.85\,m}), highlighting its superiority over FH-DRL without DRL as well as other algorithms including GDAE, CFE, and NF.

These findings underscore the robustness of the proposed algorithm, which employs DRL-based navigation to optimize path planning without relying on a static map and thus enabling efficient exploration even in cluttered environments.

\section{Conclusion}
\label{sec_conclusion}

In this work, we have proposed an integrated FH-DRL architecture that unites heuristic optimization with DRL, substantially enhancing exploration efficiency. Across three distinct experimental scenarios of varying complexity, FH-DRL integrated with DRL consistently achieved the highest exploration rates and shortest completion times, outperforming both classical and advanced frontier-based methods. This outcome demonstrates the potential of combining heuristic approaches with DRL to optimize autonomous navigation, establishing FH-DRL as a promising solution for robotic exploration tasks.

The FH-DRL framework is highly scalable and adaptable, offering robust exploration solutions for applications in autonomous driving, industrial robotics, and space exploration. Empirical evidence confirms FH-DRL as an effective, innovative approach for a range of industrial settings. Future research efforts will focus on refining heuristic functions and incorporating additional sensor modalities to further strengthen robustness and adaptability.

\section*{Acknowledgment}
\AckMiddleSizedResearcherUAM
\AckKIAT

\printbibliography

\begin{IEEEbiography}[{\includegraphics[width=1in,height=1.25in,clip,keepaspectratio]{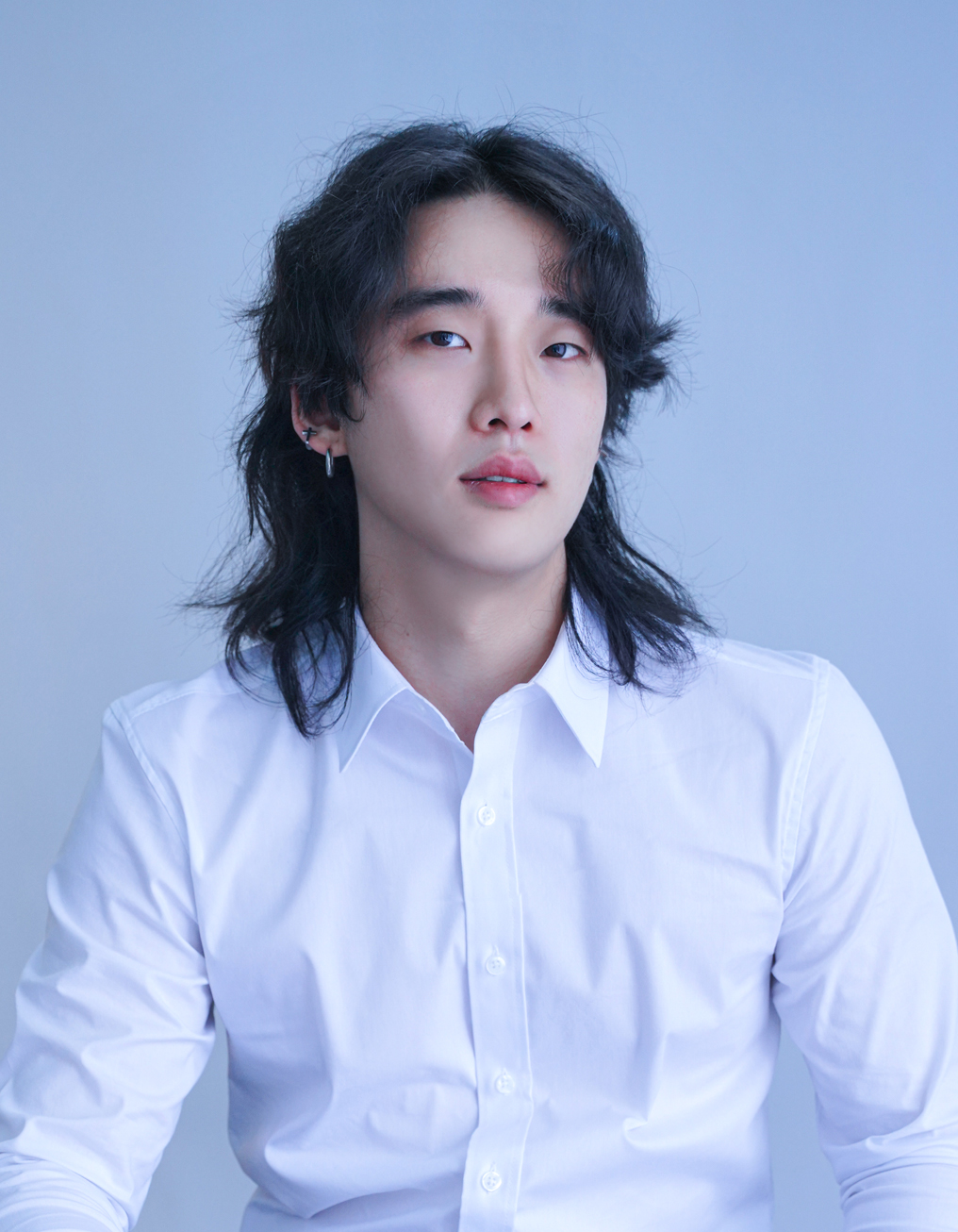}}]{Seunghyeop Nam} is a undergraduate researcher of computer science in Konkuk University Seoul Korea. He is in Distributed Multimedia Systems Laboratory (DMS Lab) at Konkuk University since 2021. His interests of research are deep reinforcement learning, robotic mechatronics, sensor fusion, vision deep learning and path planning of mobile robot.
\end{IEEEbiography}

\begin{IEEEbiography}[{\includegraphics[width=1in,height=1.25in,clip,keepaspectratio]{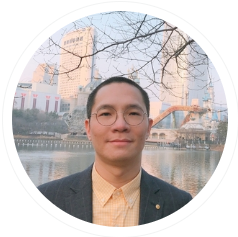}}]{Tuan Anh Nguyen} (Ph.D.'15, M.Sc.'10, B.Eng.'08) is an Academic Research Professor at Konkuk University's Aerospace Design-Airworthiness Institute in Seoul, South Korea. He is a member of IEEE, IEEE Computer, Robotics and Automation (IEEE RAS), Aerospace and Electronic Systems (IEEE AESS), and Reliability Societies. He earned his Ph.D. in Computer Science and Systems Engineering from Korea Aerospace University and his MSc and BEng in Mechatronics from Hanoi University of Science and Technology. His research focuses on Dynamics and Control Theory, AI-based Digital Twin Systems, and Autonomous Intelligent Systems.
\end{IEEEbiography}

\begin{IEEEbiography}[{\includegraphics[width=1in,height=1.25in,clip,keepaspectratio]{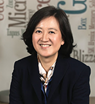}}]{Eunmi Choi}  currently at Kookmin University, Korea, specialises in big data infrastructure, cloud computing, intelligent systems, information security, parallel and distributed systems, and software architecture. She earned her M.S. and Ph.D. in Computer Science from Michigan State University (1991, 1997) and her B.S. from Korea University (1988). Previously, she was an assistant professor at Handong University (1998-2004). She leads the Distributed Information System and Cloud Computing Lab at Kookmin University.
\end{IEEEbiography}

\begin{IEEEbiography}[{\includegraphics[width=1in,height=1.25in,clip,keepaspectratio]{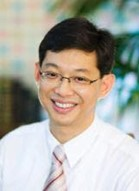}}]{Dugki Min}  received a B.S. degree in industrial engineering from Korea University in 1986, an M.S. degree in 1991 and a Ph.D. degree in 1995, both in computer science from Michigan State University. He is currently a Professor in Department of Computer Science and Engineering at Konkuk University. His research interests include cloud computing, distributed and parallel processing, big data processing, intelligent processing, software architecture, and modelling and simulation.
\end{IEEEbiography}

\vfill

\end{document}